\documentclass[sigconf,nonacm]{acmart}
\usepackage{graphicx}
\usepackage{amsmath}
\usepackage{algorithm}
\usepackage{algorithmic}
\usepackage{booktabs}
\usepackage{adjustbox}
\usepackage{color,soul}
\usepackage{subcaption}

\DeclareMathOperator*{\argmax}{argmax}

\acmYear{}
\acmVolume{}
\acmNumber{}
\acmDOI{}
\acmISBN{}
\acmConference{}
\begin{document}
%

\title{Privacy Preserving and Robust Aggregation for Cross-Silo Federated Learning in Non-IID Settings}
%
%



\author{Marco Arazzi}
\affiliation{%
  \institution{University of Pavia}
  \city{Pavia}
  \country{Italy}}
\email{marco.arazzi01@universitadipavia.it}

\author{Mert Cihangiroglu}
\affiliation{%
  \institution{University of Pavia}
  \city{Pavia}
  \country{Italy}}
\email{mert.cihangiroglu01@universitadipavia.it}

\author{Antonino Nocera}
\affiliation{%
  \institution{University of Pavia}
  \city{Pavia}
  \country{Italy}}
\email{antonino.nocera@unipv.it}


\begin{abstract}
Federated Averaging remains the most widely used aggregation strategy in federated learning due to its simplicity and scalability. However, its performance degrades significantly in non-IID data settings, where client distributions are highly imbalanced or skewed. Additionally, it relies on clients transmitting metadata, specifically the number of training samples, which introduces privacy risks and may conflict with regulatory frameworks like the European GDPR. In this paper, we propose a novel aggregation strategy that addresses these challenges by introducing class-aware gradient masking. Unlike traditional approaches, our method relies solely on gradient updates, eliminating the need for any additional client metadata, thereby enhancing privacy protection. Furthermore, our approach validates and dynamically weights client contributions based on class-specific importance, ensuring robustness against non-IID distributions, convergence prevention, and backdoor attacks. Extensive experiments on benchmark datasets demonstrate that our method not only outperforms FedAvg and other widely accepted aggregation strategies in non-IID settings but also preserves model integrity in adversarial scenarios. Our results establish the effectiveness of gradient masking as a practical and secure solution for federated learning.


\end{abstract}




\maketitle

\section{Introduction}
\numberwithin{equation}{section}
Federated Learning (FL) \cite{fl-mcmahan, konecny2016federated, bonawitz2017practical, bonawitz2019towards, li2019federated} has emerged as a scalable and privacy-conscious framework for distributed machine learning, enabling collaborative model training without requiring centralized access to raw data. This decentralized approach addresses critical privacy and regulatory constraints, such as those mandated by the General Data Protection Regulation (GDPR) in Europe~\cite{gdpr}, making it particularly suitable for privacy-sensitive domains. Cross-silo federated learning\cite{huang2022crosssilofederatedlearningchallenges}, where a limited number of institutions (e.g., hospitals, banks, or research centers) collaboratively train a model, has gained attention due to its practical relevance in scenarios with high-value, privacy-sensitive datasets. In a typical FL setup, learning is organized into two primary entities: clients that train local models and a central aggregator that coordinates and aggregates updates. Each client, which could be an individual device or an institution, performs local training on its private dataset and periodically transmits model updates to the aggregator. The aggregator then aggregates these contributions, typically using weighted averaging~\cite{fl-mcmahan}, to produce a new global model, which is subsequently distributed back to the clients. This iterative process continues until convergence.

Despite these advantages, FL faces fundamental challenges when applied to real-world non-IID (non-independent and identically distributed) settings, where client data distributions exhibit significant statistical heterogeneity. In non-IID scenarios, the assumption that each client’s local dataset follows the same underlying distribution as the global dataset no longer holds. This heterogeneity can manifest in various forms, including label distribution skew, where clients have access to only a subset of classes; feature distribution skew, where input distributions vary across clients due to domain-specific factors; and quantity skew, where clients possess differing amounts of training data. Cross-silo FL, particularly when dealing with image datasets, often encounters these challenges, leading to slow convergence and degraded model performance \cite{zhao2018federated}. Traditional aggregation methods, such as Federated Averaging (FedAvg) \cite{fl-mcmahan}, assume relatively uniform data distributions across clients, making them prone to model bias and overfitting to dominant client patterns in non-IID environments. A comprehensive study \cite{li2022federated} indicates that classification accuracy in FL can degrade by 10–20\% under non-IID scenarios compared to IID setups.

Personalized Federated Learning (PFL) has emerged as a key strategy to tackle heterogeneity in federated learning by enabling clients to maintain personalized models rather than converging to a single global model. Several PFL approaches, including pFedMe \cite{dinh2022personalizedfederatedlearningmoreau}, Ditto \cite{li2021dittofairrobustfederated}, and Clustered Federated Learning (CFL) \cite{sattler2019clusteredfederatedlearningmodelagnostic}, enhance performance by tailoring model updates to client-specific distributions. Some techniques employ regularization to balance local and global objectives \cite{dinh2022personalizedfederatedlearningmoreau}, while others leverage clustering to group clients based on data similarity \cite{sattler2019clusteredfederatedlearningmodelagnostic}. More recently, FedAMP \cite{huang2021personalizedcrosssilofederatedlearning} introduced attentive message passing to facilitate pairwise collaboration between clients with similar data. Despite their effectiveness, these methods rely on metadata exchange, including client similarity scores, statistical summaries, or model distances, which introduces privacy risks. Such metadata can be exploited by adversaries to infer sensitive client attributes, potentially violating GDPR’s data minimization and purpose limitation principles \cite{gdpr}.

In particular, most of the existing aggregation strategies requires the exchange of metadata containing client dataset sizes or statistical summaries, for weighting updates. While this information aids optimization, recent research \cite{10555053} has shown that sharing such information can lead to privacy risks, as adversaries may infer sensitive attributes about clients. For example, dataset size metadata may inadvertently reveal institutional characteristics, such as the scale of operations or population demographics. The risks of metadata leakage are further amplified under adversarial settings, where malicious clients can exploit this information to infer vulnerabilities or manipulate aggregation mechanisms.  

To mitigate metadata leakage, privacy-preserving techniques such as secure multi-party computation and homomorphic encryption have been proposed. While these techniques offer strong theoretical guarantees, their reliance on cryptographic operations introduces high computational and communication costs, making them impractical for many real-world cross-silo FL applications, especially when operating under constrained bandwidth or resource limitations~\cite{liu2022privacy, stripelis2022semi, zhang2020batchcrypt}. Furthermore, these techniques primarily address privacy concerns but fail to mitigate the negative effects of non-IID data on convergence and remain vulnerable to adversarial manipulations, such as Byzantine faults and backdoor attacks~\cite{xie2019dba, mei2023privacy}. These attacks exploit the aggregation mechanisms in FL to either prevent model convergence or implant malicious behaviors into global models, ultimately compromising model reliability and security.

In this paper, we propose a novel and effective aggregation method based on class-aware gradient masking to address these challenges. Our method eliminates reliance on client metadata, preserves privacy, and enhances robustness against Byzantine faults and backdoor attacks, ensuring better performance in heterogeneous federated learning environments. Specifically, we design an adaptive masking mechanism that prioritizes gradients relevant to each class while selectively attenuating potential adversarial updates. By applying this approach to image datasets under highly non-IID conditions, we demonstrate significant improvements in both accuracy and attack resilience. Our main contributions are as follows:

\begin{itemize}  
    \item We design an adaptive masking mechanism that by design defends against backdoor attacks and convergence prevention attacks, demonstrating improved robustness while preserving privacy through gradient-only aggregation.  
    
    \item We evaluate the proposed method under highly challenging non-IID settings using Dirichlet distributions ($\alpha = 0.125, 0.3, 0.5$) and validate its performance across multiple datasets and attack scenarios.  
\end{itemize}

Extensive experiments demonstrate that our method consistently outperforms baseline approaches such as FedAvg, FedProx, FedNova, and SCAFFOLD, achieving 5–20\% higher accuracy on average across various datasets and Dirichlet distributions. Furthermore, our approach significantly mitigates adversarial threats, substantially lowering attack success rates while maintaining robust model performance.

\section{Preliminaries}
\label{preliminaries}
In this section, we first formally define federated learning and its key components. We then discuss various aggregation strategies, including FedAvg\cite{fl-mcmahan}, FedProx\cite{FedProx}, FedNova\cite{FedNova}, and SCAFFOLD\cite{scaffold}, analyzing their strengths, limitations, and privacy implications. These aggregation methods play a crucial role in determining the efficiency, convergence behavior, and privacy guarantees of federated learning systems, making their evaluation essential for privacy-preserving FL.

\subsection{\textbf{Federated Learning}}
FL is a decentralized machine learning paradigm that enables multiple clients to collaboratively train a global model (GM) while keeping their data localized. Given a set of $N$ clients with local datasets $\{D_1, D_2, \dots, D_N\}$, the goal is to minimize a global objective function:

\begin{equation}
\min_{w} F(w) = \sum_{i=1}^{N} \frac{n_i}{n} F_i(w)
\end{equation}

where:  
\begin{itemize}
    \item $w$ represents the model parameters,  
    \item $F_i(w)$ is the local loss function for client $i$,  
    \item $n_i$ is the number of data samples at client $i$, and  
    \item $n = \sum_{i=1}^{N} n_i$ is the total number of data samples across all clients.  
\end{itemize}
Each client computes updates based on its local data and sends these updates to a central server. The server aggregates the updates to optimize the global model. The most common aggregation method, FedAvg, combines model updates weighted by the size of each client’s dataset:

\begin{equation}
w_{t+1} = \sum_{i=1}^{N} \frac{n_i}{n} w_i^t
\end{equation}
where $w_i^t$ denotes the local model update from client $i$ at round $t$. 

While federated learning provides a framework for distributed training, it faces two major challenges. First, its basic aggregation strategy struggles in non-IID data distributions, which are very common in real-world scenarios, leading to poor convergence and suboptimal performance. Second, it does not inherently defend against attacks, including convergence prevention attacks and backdoor attacks, leaving the global model vulnerable to malicious updates. In the following section, we discuss existing aggregation methods designed to enhance performance in non-IID data distributions.

\subsection{\textbf{Learning and Aggregation strategies for Federated Learning}}

Several aggregation strategies have been proposed to address the challenges of federated learning in non-IID environments. Among them, FedProx\cite{FedProx}, FedNova\cite{FedNova}, and SCAFFOLD~\cite{scaffold} are widely recognized, each introducing unique mechanisms that influence both local training and global model aggregation to improve performance and convergence.

\subsubsection{\textbf{FedProx}}  
It builds on FedAvg by adding a proximal term to the loss function. This term penalizes large deviations from the global model, addressing client drift caused by heterogeneous data distributions. The objective function is defined as:

\begin{equation}
F_i(w) = L_i(w) + \frac{\mu}{2} \|w - w^t\|^2 
\end{equation}

where:
\begin{itemize}
    \item $L_i(w)$ is the local loss function for client $i$,
    \item $w$ represents the model parameters,
    \item $w^t$ is the global model at round $t$, and
    \item $\mu$ is a hyperparameter controlling the strength of the proximal term.
\end{itemize}
This approach claims to improve stability during training by reducing the impact of client drift, particularly in non-IID scenarios. However, FedProx still relies on metadata exchange, including dataset sizes, to perform weighted aggregation. While FedProx modifies local training dynamics, it inherits the standard FedAvg aggregation strategy, where client updates are typically weighted by dataset sizes. This means that, although the proximal term itself does not introduce metadata dependencies, the global aggregation step still requires dataset size information, exposing clients to potential metadata leakage.

\subsubsection{\textbf{FedNova}}  
This strategy addresses imbalances in client contributions by normalizing updates based on local optimization steps. Instead of relying solely on dataset size, it scales updates to ensure fairness across varying workloads and data distributions. The aggregation step is given by:

\begin{equation}
w_{t+1} = w_t + \frac{\sum_{i=1}^{N} p_i \Delta w_i^t}{\sum_{i=1}^{N} p_i} 
\end{equation}

where:
\begin{itemize}
    \item $\Delta w_i^t$ represents the model update from client $i$ at round $t$,
    \item $p_i = \frac{n_i \tau_i}{\sum_{j=1}^{N} n_j \tau_j}$ is a normalized weighting factor,
    \item $\tau_i$ is the number of local training steps performed by client $i$.
\end{itemize}

By incorporating both $n_i$ (dataset size) and $\tau_i$ (local training steps) into the aggregation process, FedNova increases reliance on metadata compared to standard FedAvg. While this normalization ensures more balanced updates and improves convergence in non-IID settings, it introduces additional privacy risks.

\subsubsection{\textbf{SCAFFOLD}}  
It employs a variance-reduction mechanism by maintaining control variates at both server and client levels. These control variates correct client drift caused by non-IID data, reducing gradient variance and improving convergence rates. The update rule for each client is defined as:

\begin{equation}
w_i^{t+1} = w_i^t - \eta \left( \nabla F_i(w_i^t) - c_i + c \right) 
\end{equation}

where:
\begin{itemize}
    \item $\eta$ is the learning rate,
    \item $c_i$ and $c$ are the control variates for the client and server, respectively.
\end{itemize}
The server aggregates updates and adjusts the control variates to minimize the variance in gradients:

\begin{equation}
c = c + \frac{1}{N} \sum_{i=1}^{N} \left( \nabla F_i(w_i^t) - \nabla F_i(w^t) \right) 
\end{equation}
This method aims to improve convergence rates under non-IID conditions by reducing gradient discrepancies. While SCAFFOLD is designed to enhance learning stability, it requires the exchange of control variates ($c_i$ and $c$) between clients and the server, which can introduce privacy concerns. These control variates, designed to correct client drift, may inadvertently encode statistical properties of local datasets, potentially revealing sensitive information. SCAFFOLD’s reliance on this additional metadata could make clients more distinguishable, increasing the risk of targeted attacks in adversarial settings. For a detailed discussion on the potential vulnerabilities associated with control variate exchanges in federated learning, refer to the study on backdoor attacks against SCAFFOLD \cite{han2024badsflbackdoorattackscaffold}.

\subsubsection{\textbf{Limitations and Observations}}  

In cross-silo federated learning, where a small number of institutions (e.g., hospitals, banks, or research centers) collaborate, aggregation strategies must ensure both efficient learning and privacy protection despite significant statistical heterogeneity among clients. However, prior work~\cite{li2022federated} has shown that existing aggregation methods often fail to outperform FedAvg in real-world settings, primarily due to convergence inefficiencies and fairness concerns. FedAvg, while simple and widely used, struggles with client drift in cross-silo scenarios where institutions have highly non-IID data distributions, leading to slower convergence and potential model bias\cite{fl-mcmahan}. FedProx attempts to mitigate this by introducing a proximal term that constrains local updates\cite{FedProx}, but it does not modify the aggregation mechanism itself, meaning that it still relies on dataset size-weighted averaging, which can disproportionately favor institutions with larger datasets and overlook smaller contributors.

Beyond performance issues, security vulnerabilities remain a major concern. Most existing methods, including FedAvg and FedProx, lack built-in defenses against adversarial attacks such as convergence manipulation and backdoor poisoning, making them unsuitable for security-sensitive applications. Moreover, aggregation mechanisms like FedNova and SCAFFOLD require explicit metadata exchange, such as dataset sizes, local step counts, or control variates, which can leak statistical information about client data~\cite{han2024badsflbackdoorattackscaffold}. This metadata leakage increases the risk of client tracking and adversarial exploitation, allowing an attacker to infer participating institutions’ data distributions or even identify vulnerable clients. In cross-silo settings, where clients may represent high-value organizations with sensitive data, such risks become even more critical.

To address these challenges, we propose a privacy-preserving aggregation strategy that simultaneously improves convergence efficiency, mitigates metadata leakage, and enhances robustness against adversarial threats in federated learning under non-IID conditions.

\begin{algorithm}
\caption{Adaptive Masked Aggregation with Class-Specific Evaluation and Mask Averaging}
\label{alg:adaptive_masked_aggregation}
\begin{algorithmic}[1]

\STATE \textbf{Input:} Local models $\{LM_i^r\}_{i=1}^{N}$, masks $\{M_i^{r-1}\}_{i=1}^{N}$, validation sets $\{\mathcal{V}_c\}_{c=1}^{C}$, gradient threshold $\tau$, scale-down factor $\gamma$, mask update factor $\beta$, and zip percentage $p$.

\STATE \textbf{Output:} Updated global model $GM^R$.

\FOR{each round $r = 1$ to $R$}

    \STATE \textbf{Step 1: Class Assignment and Mask Generation}
    \STATE Select mask generation method:
    \begin{itemize}
        \item \textbf{Method 1:} For each class $c \in \{1, \dots, C\}$, evaluate all $\{LM_i^r\}$ on $\mathcal{V}_c$ and select top models for each class.
        \item \textbf{Method 2:} For each client $i$, assign $c^*_i = \arg\max_{c} \text{Accuracy}(LM_i^r, \mathcal{V}_c)$.
    \end{itemize}

    \FOR{each client $i$}
        \STATE Compute gradients $\nabla \mathcal{L}(LM_i^r, \mathcal{V}_{c^*_i})$.
        \FOR{each parameter $p$ in $LM_i^r$}
            \STATE Compute gradient magnitude $g_p = |\nabla \mathcal{L}(p)|$.
            \STATE Generate mask $M_i^{\text{new}}(p)$ using $f(\cdot)$:
            \[
            M_i^{\text{new}}(p) =
            \begin{cases} 
            1 & \text{if } g_p > \tau, \\
            \gamma & \text{otherwise.}
            \end{cases}
            \]
        \ENDFOR
        \STATE Update mask: $M_i^r = (1 - \beta) M_i^{\text{new}} + \beta M_i^{r-1}$.
    \ENDFOR

    \STATE \textbf{Step 2: Weighted Aggregation}
    \STATE Compute importance weights $w_i^r = \frac{\sum_{p} M_i^r(p)}{\sum_{j=1}^{N} \sum_{p} M_j^r(p)}$.
    \FOR{each parameter $p$ in $GM^r$}
        \STATE Aggregate: $GM^r(p) = \sum_{i=1}^{N} w_i^r \cdot M_i^r(p) \odot LM_i^r(p)$.
    \ENDFOR

    \STATE \textbf{Step 3: Update Global Model}
    \STATE $GM^{r+1} \leftarrow GM^r$.
\ENDFOR
\STATE \textbf{Return:} Final global model $GM^R$.

\end{algorithmic}
\end{algorithm}
\section{Methodology}

\begin{figure}[t]
    \centering
    \includegraphics[width=\linewidth]{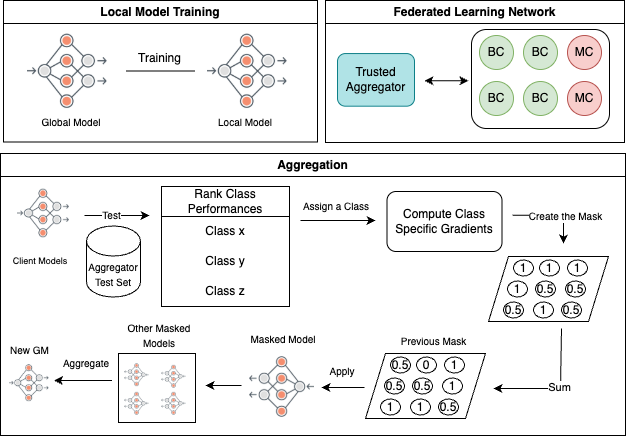}
    \caption{An overview of the workflow for our proposed approach within a federated learning network. Local clients train the current global model using their own data, after which the updates are sent to the trusted aggregator. The participating clients can be either benign (\textbf{BC}) or malicious (\textbf{MC}). The process begins with the aggregator evaluating client models using its test set to determine their performance across different classes. Based on this evaluation, the aggregator assigns a dominant class to each model and computes gradients relative to the assigned class. A mask is then generated from these gradients and updated by summing it with the mask from the previous round. The updated mask is applied to the model, filtering out less relevant parameters. Finally, the aggregator performs weighted averaging over the masked models, aggregating contributions from both benign and potentially malicious clients, to generate the new global model.}
    \label{fig:overview-workflow}
\end{figure}

As explained in Section-\ref{preliminaries}, existing federated learning aggregation strategies frequently rely on additional metadata beyond gradient updates to adjust for non-IID data distributions. FedAvg, the most commonly used baseline, applies dataset size-weighted averaging, which can expose sensitive information about client datasets. FedProx inherits this reliance while introducing a proximal term to stabilize local updates, but it still requires dataset size awareness to maintain training stability~\cite{FedProx}. FedNova further increases metadata dependency by incorporating both dataset sizes and local step counts, making client profiling based on computational capacity possible~\cite{FedNova}. SCAFFOLD tries to improve variance reduction through control variates, which encode statistical information about client updates and, over multiple rounds, can reveal characteristics of local data distributions~\cite{han2024badsflbackdoorattackscaffold}. These metadata dependencies introduce serious privacy risks, as adversaries can exploit dataset sizes, computational patterns, or update statistics to infer client identities, data characteristics, or even participation trends. Addressing these concerns, we propose a privacy-preserving aggregation strategy that entirely eliminates metadata reliance, requiring only gradient updates for aggregation. By removing the need for auxiliary client information, our method enhances both privacy protection and Byzantine resilience, ensuring robust learning in non-IID federated environments. Our approach is based on the observation that, in non-IID settings, client models naturally develop class-specific expertise due to data imbalances. Instead of relying on metadata-based adjustments, we introduce a class-aware masking mechanism, where the aggregator evaluates all client models across different classes and assigns a dominant class label to each client. This allows the aggregator to generate class-specific masks, highlighting important gradient updates related to each client’s dominant class. To ensure both stability and adaptability, the mask generation process is iterative, smoothing updates by averaging newly generated masks with those from previous rounds. The masked updates are then aggregated using a weighted scheme, where importance values derived from masks determine the contribution of each client’s model. By focusing only on relevant gradient updates, this method not only enhances performance in non-IID settings but also reduces vulnerabilities to backdoor attacks and convergence prevention by suppressing irrelevant or manipulated updates. A visual overview of the proposed end-to-end process, from model evaluation and mask generation to final aggregation, is provided in Figure~\ref{fig:overview-workflow}. The following subsections detail each step, from global model initialization to mask-based weighted aggregation, including the necessary formulations and design choices. A summary of all notations used throughout this section is provided in Table~\ref{tab:notations}.

\begin{table}[t]
\centering
\caption{Summary of Notations}
\label{tab:notations}
\resizebox{\columnwidth}{!}{
\begin{tabular}{ll}
\toprule
\textbf{Notation} & \textbf{Description} \\ 
\midrule
$N$       & Total number of clients \\
$A$       & Aggregator \\
$C$       & Total number of classes \\
$GM^r$    & Global model after communication round $r$ \\
$LM_i^r$  & Local model of the $i$-th client at round $r$ \\
$c^*_i$   & Dominant class for client $i$ based on accuracy \\
$\mathcal{V}$ & Validation dataset held by the aggregator \\
$\mathcal{V}_c$ & Class-specific validation data for class $c$ \\
$M_i^r$   & Mask generated for client $i$ at round $r$ \\
$M_i^{r-1}$ & Mask from the previous round ($r - 1$) \\
$w_i^r$   & Mask importance weight for client $i$ in round $r$ \\
$\mathcal{L}$ & Loss function used for gradient computation \\
$f(\cdot)$ & Thresholding function for mask generation \\
$\odot$   & Element-wise multiplication operator \\
$\alpha$  & Dirichlet parameter for data distribution \\
$\gamma$  & Scale-down factor for pruned gradients \\
$p$       & Zip percentage (percentage of weights retained) \\
$\tau$    & Gradient threshold for pruning \\
$\nabla$  & Gradient computation operator \\
$\mu$     & Regularization parameter for FedProx \\
\(\beta\) & Mask update factor balancing stability and adaptability. \\
\( \theta \) & An individual mask parameter in the mask $M_i^r$  \\
\bottomrule
\end{tabular}
}
\end{table}

 
\subsection{Global Model Initialization}

To formalize the proposed approach, we consider a federated learning setup with $N$ clients and $C$ classes. Let $GM^0$ represent the initial global model at round $r=0$. Each client $i \in {1, 2, \dots, N}$ maintains a local model $LM_i^r$ and trains it using its private local dataset $\mathcal{D}_i$. The training process continues for multiple communication rounds $r = 1, 2, \dots, R$. At round $r = 0$, the server initializes the global model:

\begin{equation}
GM^0 \leftarrow \text{Initialize randomly ( or pre-trained weights.)}
\end{equation}
This model is broadcast to all clients for local training.

\subsection{Local Model Training}
Each client updates its local model using stochastic gradient descent (SGD) on its private dataset $\mathcal{D}_i$. Let $\mathcal{L}$ represent the local loss function. The client optimizes the following objective:

\begin{equation}
\min_{w} \mathcal{L}(LM_i^r; \mathcal{D}_i).
\end{equation}
After training, the updated local model $LM_i^r$ is sent back to the server. Unlike traditional aggregation methods, which often require additional parameters, such as client dataset sizes or algorithm-specific control variables, our approach exclusively utilizes the updates from each client’s model, ensuring that privacy-related information remains undisclosed.

\subsection{Model Evaluation and Class Assignment}
\label{sec:class_assignment}
After receiving locally trained models from clients, the central aggregator evaluates their performance using a validation dataset $\mathcal{V}$. The goal is to determine the dominant class $c^*_i$ for each model $M_i$, using class-specific accuracy metrics. The dominant class is identified as the class for which a model achieves the highest classification accuracy:

\begin{equation}
c^*i = \argmax{c \in {1, \dots, C}} \mathrm{Accuracy}(M_i, \mathcal{V}_c),
\end{equation}

where $\mathcal{V}_c$ represents the subset of validation data corresponding to class $c$. This approach allows multiple models to be assigned to the same class, ensuring that knowledge from different clients contributes to the most relevant categories. The selected models then undergo gradient-based mask generation, where key parameters are preserved, and less critical updates are scaled down. This process refines model aggregation by emphasizing essential features while mitigating the influence of noisy or uninformative updates. Further details on mask generation are provided in Section~\ref{sec:mask_generation}.

\subsection{Mask Generation}
\label{sec:mask_generation}
In our proposed method, mask generation serves as a critical step to selectively preserve model parameters that are most relevant to learning, while reducing the influence of less significant updates. At every round, this process is guided by gradient magnitudes, which indicate the importance of each parameter based on its contribution to the loss function. To control the behavior of this selective preservation, the trusted aggregator employs two hyperparameters that govern the mask generation process:  

\begin{itemize}
    \item \textit{Zip Percent (\(p\))}: Specifies the proportion of parameters to retain based on their gradient magnitudes. Higher values preserve more parameters, leading to denser masks, while lower values emphasize sparsity by focusing only on the most influential gradients.  
    \item \textit{Scale Down Factor (\(\gamma\))}: Determines the scaling applied to parameters that fall below the threshold set by \(p\). Instead of completely discarding less important weights, \(\gamma\) scales them down to prevent abrupt changes and maintain numerical stability.  
\end{itemize}
Both $\gamma$ and $p$ take values between 0 and 1, allowing fine-tuned control over pruning severity and sparsity.

\subsubsection{\textbf{Gradient-Based Mask Generation}}After identifying the dominant class $c^*_i$ for each client $i$ at round $r$ (as defined in Section~\ref{sec:class_assignment}), masks are generated by analyzing gradient information derived from the model’s loss function. Specifically, we compute the gradients of the loss function $\mathcal{L}$ with respect to the model parameters, using the validation data associated with the dominant class $\mathcal{V}_{c^*_i}$:

\begin{equation}
G_i^r = \nabla \mathcal{L}(LM_i^r, \mathcal{V}_{c^*_i}),
\end{equation}
where $G_i^r$ represents the computed gradients for client $i$ at round $r$. Next, we process these gradients to generate the mask $M_i^r$. A thresholding function, denoted as \( f_p(\cdot) \), is applied to generate a selection mask that determines whether a model weight is retained or attenuated. Each weight is assigned either a value of \(1\) (indicating that the weight is kept unchanged) or the Scale-Down Factor (\(\gamma\)) (indicating that the weight is reduced to mitigate its influence). This process identifies and highlights important parameters by analyzing their contribution to the dominant class-specific loss. The thresholding function is controlled by hyperparameter \(p\), which specifies the proportion of parameters to retain. Formally, the mask \( M_i^r \) for client \( i \) at round \( r \) is computed as:

\begin{equation}
M_i^r = f_p(G_i^r),
\end{equation}
where \( f_p(\cdot) \) represents the thresholding function controlled by \( p \) (\textit{zip percent}), and \( G_i^r \) represents the gradient magnitudes of the model parameters with respect to the dominant class for client $i$ at round $r$. Parameters are ranked based on their magnitudes, and the top \( p\% \)  are selected as important, forming the mask \( M_i^r \). This process produces a mask that distinguishes critical parameters, enabling the preservation of high-importance updates while allowing the suppression of less significant gradients in subsequent steps. The detail explanation of \( f_p(\cdot) \) is explained in the following section-\ref{subsec:thresholding_pruning}

 \subsubsection{\textbf{Thresholding and Pruning with Zip Percent}}\label{subsec:thresholding_pruning}The thresholding and pruning process is designed to identify and retain the most important parameters based on their gradient magnitudes. It operates in two key steps, controlled by the \textit{zip percent} (\(p\)) and \textit{scale down factor} (\(\gamma\)) hyperparameters.

\begin{enumerate}
    \item \textit{Threshold Selection:}  
    Given a gradient tensor \( G_i^r \), the top \( p\% \) of parameters with the largest absolute magnitudes are retained. The threshold $\tau_i^r$ for client $i$ at round $r$ is computed dynamically based on the selected percentile:

    \begin{equation}
    \tau_i^r= \text{top-k}\left(|G_i^r|, k = p \cdot \text{size}(G_i^r)\right),
    \end{equation}

    where \( |G_i^r| \) denotes the absolute gradient values of client $i$ at round $r$ , and \( k \) represents the number of parameters to retain based on \(p\).

    \item \textit{Pruning and Mask Creation:}  
    Using the computed threshold $\tau_i^r$ for client $i$ at round $r$ , the gradients are evaluated, and a  mask \( M_i^r \) is generated. Parameters exceeding the threshold are preserved, while the remaining parameters are scaled down using the \textit{scale down factor} (\(\gamma\)) to prevent abrupt updates and ensure numerical stability:
    \begin{equation}
     M_i^r =
    \begin{cases} 
    1 & \text{if } |G_i^r| > \tau, \\
    \gamma & \text{otherwise.}
    \end{cases}
    \end{equation}

\end{enumerate}
This process ensures that the most relevant parameters are prioritized during aggregation, enabling the model to focus on critical updates while mitigating noise from less significant gradients. The hyperparameters \( p \) and \( \gamma \) provide additional flexibility to balance sparsity and stability in mask generation.

\subsubsection{\textbf{Incremental Updates for Stability}}  

To smooth transitions and prevent abrupt fluctuations between rounds, masks are incrementally updated by averaging them with masks from the previous round. Formally, the updated mask \(M_i^r\) for client \(i\) at round \(r\) is computed as:

\begin{equation}
M_i^r = (1 - \beta) M_i^r + \beta M_i^{r-1},
\end{equation}
where \(M_i^{r-1}\) represents the mask from the previous round, and \(\beta \in [0, 1]\) is the mask update factor. This hyperparameter controls the balance between stability and responsiveness:
\begin{itemize}
    \item A higher \(\beta\) gives greater weight to the previous mask, reinforcing stability and reducing fluctuations.
    \item A lower \(\beta\) emphasizes the current mask, enhancing adaptability to dynamic changes in model parameters.
\end{itemize}
In our experiments, \(\beta = 0.4\) strike a balance between two extremes (0 and 1), ensuring smooth transitions without sacrificing responsiveness to parameter updates.

\subsubsection{\textbf{Final Mask Application}}  

The computed masks are applied element-wise to the local model parameters:

\begin{equation}
\tilde{LM}_i^r = LM_i^r \odot M_i^r,
\end{equation} 
where $\odot$ represents element-wise multiplication. This masking operation selectively attenuates unimportant weights, ensuring that updates focus on the most informative parameters. 

\subsection{\textbf{Weighted Aggregation}}  


After generating masks, the masked local models are aggregated using a \textbf{weighted averaging scheme} to form the updated global model. 
The \textit{weights are derived from the importance of each mask}, ensuring that models with a higher number of \textit{retained parameters} contribute more to the global update. The global model at round \( r \) is computed as:

\begin{equation}
GM^{r+1} = \frac{\sum_{i=1}^N \omega_i^r \cdot \left( LM_i^r \odot M_i^r \right)}{\sum_{i=1}^N \omega_i^r},
\end{equation}

where:
\begin{itemize}
    \item \( \omega_i^r \) is the importance score for client \( i \) at round \( r \).  
    \item \( M_i^r \) represents the mask generated for client \( i \).  
    \item \( \odot \) denotes element-wise multiplication, ensuring that only the retained parameters (based on the mask) are considered during aggregation.
\end{itemize}
The importance score \( \omega_i^r \) is computed as the \textit{sum of the mask values}, indicating the total number of preserved parameters:

\begin{equation}
\omega_i^r = \sum_{\theta \in M_i^r} M_i^r(\theta),
\end{equation}
where:
\begin{itemize}
    \item \( \theta \) represents an \textit{individual mask parameter}, corresponding to a specific position in the mask \( M_i^r \).
    \item \( M_i^r(\theta) \) is the \textit{mask value} at position \( \theta \).
\end{itemize}

By assigning weights based on \textit{the total retained parameters}, this approach ensures that clients contributing more \textit{relevant updates} have a \textit{stronger impact} on the aggregated global model.

\section{Attack Model}
\label{sec:attack_model}

In our attack model, we focus on two primary types of attacks: convergence prevention (CP) and backdoor attacks (BA). These attack types are widely recognized as significant threats in federated learning systems due to their potential to compromise both performance and security. CP aims to hinder the training process by injecting malicious updates, causing the global model to either fail in achieving optimal performance or diverge entirely. To evaluate the resilience of our approach under such attacks, we simulate data poisoning as the attack mechanism. In our non-IID setting, we design the attack by randomly selecting samples from various classes and flipping their labels randomly. This disrupts the model’s learning process by introducing mislabeled data, preventing effective convergence. This setup reflects a challenging scenario for the global model, simulating a worst-case scenario for model stability.  

For BA, we implement a distributed backdoor attack (DBA), where adversaries inject hidden vulnerabilities into the global model. A successful backdoor attack must not affect the final clean accuracy of the model. If the model's overall accuracy drops significantly, the attack is easily detectable and ineffective. At the same time, the attack success rate (ASR) must be high, ensuring that the backdoor reliably triggers misclassification when presented with a specific pattern or input. To evaluate this, our implementation of DBA uses four distinct triggers distributed across malicious clients, creating a more realistic and challenging adversarial environment. This setup ensures that the attack remains undetected while maintaining its effectiveness, highlighting the critical security risks posed by such attacks in federated learning systems. It is important to note that in each of these setups, the majority of clients cannot be attackers. To ensure this, we adhere to the constraint:
\begin{equation}
m \leq \frac{N - 1}{2},
\end{equation}
where \( m \) is the number of malicious clients and \( N \) is the total number of clients. This ensures that the majority of updates remain benign, reflecting a realistic and practical adversarial scenario. These two attack types epresent fundamental threats that exploit different weaknesses in federated learning systems, namely model stability and prediction integrity. Furthermore, they are frequently studied in the literature as benchmarks for evaluating the robustness of federated learning systems~\cite{sun2019reallybackdoorfederatedlearning,bhagoji2019analyzing, marco_arazzi_evading_model_poisoning, Shejwalkar2021ManipulatingTB}. By addressing these scenarios, we provide comprehensive insights into the resilience of our approach against both targeted and untargeted adversarial strategies.

\section{Experiments}
This section describes the experiments conducted to evaluate our proposed method and compare it against existing approaches. We begin by introducing the datasets used, followed by a discussion of the model architecture, attack scenarios, baseline methods, and evaluation metrics. We then present the experimental results, analyzing the performance of our approach under various conditions. Finally, we conclude the section with key observations and insights derived from the experiments.

\subsection{Experimental Setup}
In this section, we provide a detailed description of the experimental settings used in our study.
\subsubsection{\textbf{Datasets.}}
We evaluate our approach using three datasets: CIFAR10 \cite{Krizhevsky2009LearningML}, CIFAR100 \cite{Krizhevsky2009LearningML}, and FashionMNIST \cite{xiao2017fashionmnistnovelimagedataset}. We selected these datasets inspired by their widespread use in prior studies, ensuring comparability and relevance to established benchmarks in federated learning research~\cite{xie2019dba, walter2024mitigating, Li_2022, reyes2021precisionweightedfederatedlearning, Konecn2016FederatedLS}. CIFAR10 contains 60,000 32x32 RGB images across 10 classes, with 50,000 training and 10,000 test images. CIFAR100 follows the same structure but includes 100 classes. FashionMNIST is a dataset of 70,000 28x28 grayscale images divided into 10 classes, with 60,000 images used for training and 10,000 for testing. All datasets simulate non-IID settings using a Dirichlet distribution with concentration parameters $\alpha = 0.125$, $\alpha = 0.3$, and $\alpha = 0.5$ consistent with previous studies~\cite{li2022federated}. Data is distributed among 10 clients.

\subsubsection{\textbf{Model architecture.}} 
We use ResNet-18 as the base model with randomly initialized weights. Each client trains the model locally using Stochastic Gradient Descent (SGD) with a learning rate of 0.01, momentum of 0.9, weight decay of 1e-4, and a batch size of 64. Local training is performed for 10 epochs per round over a total of 100 federated learning rounds.

\subsubsection{\textbf{Attacks.}}
We evaluate robustness against two adversarial attacks: \textit{Distributed Backdoor Attack (DBA)} and \textit{Convergence Prevention (CP)}. To simulate a Convergence prevension attack we have implemented a data poisoning mechanism as explained in the Section-\ref{sec:attack_model}.  In the \textit{CP} attack, 49\% of the clients are malicious, and 40\% of each malicious client's data is poisoned. In the DBA, attackers embed pixel-based triggers into training samples to induce targeted misclassification. We set the malicious client percentage to 30\% and the backdoored data ratio to 20\% in the DBA setting. The CP attack instead is based on an untargeted label-flipping attack that aims at degrading the final model accuracy.

\subsubsection{\textbf{Evaluation metrics.} }
Performance is measured using Test Accuracy (A) on clean data and Attack Success Rate (ASR) for adversarial scenarios. 

\subsubsection{\textbf{Baseline Approaches}}
To evaluate the effectiveness of our proposed aggregation strategy, we compare it against widely used \textit{federated learning aggregation methods}, including \textit{nwFedAvg}, \textit{FedAvg}, \textit{FedProx}, \textit{FedNova}, and \textit{SCAFFOLD}. \textit{nwFedAvg} (non-weighted FedAvg) aggregates updates through \textit{simple averaging}, disregarding client dataset sizes, while \textit{FedAvg} performs \textit{weighted averaging} based on dataset size. This distinction allows us to assess the impact of dataset size consideration on model performance under different levels of data heterogeneity. For aggregation methods that require hyperparameter tuning, we ensure fair comparisons by adopting \textit{standard parameter configurations}, following prior work~\cite{li2022federated}. Specifically, we set the \textit{proximal term} \( \mu = 0.01 \) for \textit{FedProx}, maintain a \textit{momentum parameter} of \( 0.9 \) for \textit{FedNova}, and initialize \textit{control variates} in \textit{SCAFFOLD}, updating them throughout training to correct for client drift. To account for varying degrees of \textit{data heterogeneity}, we evaluate these methods under multiple \textit{non-IID settings}, characterized by different Dirichlet parameter values (\( \alpha = 0.125, 0.3, 0.5 \)). The corresponding client data distributions for these settings are illustrated in Figures~\ref{fig:distribution_alpha_0.125}, \ref{fig:distribution_alpha_0.3}, and \ref{fig:distribution_alpha_0.5}, providing a visual representation of how data is allocated across clients at different levels of heterogeneity.  By including both \textit{weighted and non-weighted aggregation strategies}, along with methods that introduce \textit{proximal constraints, normalization, or variance reduction}, we ensure a \textit{comprehensive comparison} that highlights the trade-offs between convergence efficiency, fairness, and robustness in federated learning.

\subsection{Experimental Results}
In this section, we evaluate our proposed aggregation strategy in federated learning under non-IID settings and compare its performance against traditional aggregation methods on CIFAR-10, CIFAR-100, and FashionMNIST. Our approach is designed to improve model convergence and accuracy in heterogeneous data distributions, addressing challenges commonly faced in federated learning. While our method is \textbf{not explicitly designed as a defense mechanism}, we also assess its behavior under adversarial settings, including Distributed Backdoor Attacks (DBA) and Convergence Prevention Attacks, to understand how it inherently responds to these threats. The results are structured as follows: In Section~\ref{sec:clean_results}, we present the performance of our method on CIFAR-10, CIFAR-100, and FashionMNIST under different levels of data heterogeneity ($\alpha = 0.125, 0.3, 0.5$), comparing it with existing aggregation techniques. In Section~\ref{sec:dba_results}, we analyze its behavior under Distributed Backdoor Attacks, highlighting how its design influences the attack success rates. Finally, in Section~\ref{sec:convergence_prevention}, we examine its response to Convergence Prevention Attacks, showcasing its ability to maintain stability even when faced with adversarial updates.

\subsubsection{\textbf{Performance Results}}
\label{sec:clean_results}
Table~\ref{tab:cleanaccuracy} presents the accuracy results for CIFAR-10, CIFAR-100, and Fashion-MNIST across different aggregation methods and varying heterogeneity levels ($\alpha = 0.125, 0.3, 0.5$). In particular, we compare our proposed method against \textit{nwFedAvg}, \textit{FedAvg}, \textit{FedProx}, \textit{FedNova}, and \textit{SCAFFOLD}, evaluating its effectiveness in handling data heterogeneity and improving model performance. Our method consistently outperforms traditional aggregation approaches, demonstrating significant improvements across all datasets and non-IID settings. As shown in Table~\ref{tab:accuracy_improvement}, for CIFAR-10, our approach achieves a substantial 14.85\% improvement over FedAvg at $\alpha = 0.125$, confirming its ability to handle extreme data heterogeneity more effectively. Even as the data distribution becomes more balanced at $\alpha = 0.3$ and $\alpha = 0.5$, our method maintains a consistent lead, surpassing FedAvg by 11\% and 6.6\%, respectively. These results highlight the robustness of our approach in both highly skewed and moderately heterogeneous federated learning settings, where traditional aggregation methods suffer from performance degradation.

For CIFAR-100, a more complex dataset with a larger number of classes, our method demonstrates even greater improvements. At $\alpha = 0.3$ and $\alpha = 0.5$, our approach surpasses FedAvg by 22.87\% and 21.73\%, respectively, significantly enhancing accuracy in challenging multi-class federated learning environments. Even under the most extreme heterogeneity setting of $\alpha = 0.125$, where standard methods experience severe accuracy drops, our method maintains a strong performance advantage, improving upon FedAvg by 5.07\%. These results further validate the adaptability of our approach in handling high-dimensional, heterogeneous data distributions.

On Fashion-MNIST, our method continues to deliver the highest accuracy across $\alpha = 0.3$ and $\alpha = 0.5$, outperforming FedAvg by 1.34\% and 1.46\%, respectively. At $\alpha = 0.125$, the performance difference is minimal, reflecting the fact that traditional methods perform competitively on simpler datasets. However, our approach still maintains a strong and stable performance across all settings, demonstrating its effectiveness in both complex and lower-dimensional datasets.

Overall, these results clearly establish the superiority of our approach over existing methods. Our model consistently achieves higher accuracy, better stability, and stronger generalization across different datasets, varying levels of heterogeneity, and multiple aggregation strategies. The improvements are particularly pronounced in highly non-IID settings, where traditional methods struggle, further reinforcing the robustness and adaptability of our method in practical federated learning applications.

\begin{table}[h!]
\centering
\caption{Performance results (accuracy) for CIFAR-10, CIFAR-100, and FashionMNIST under different non-IID settings with Dirichlet parameter values \( \alpha = 0.125, 0.3, 0.5 \), and \textit{10} clients across various aggregation methods. The hyperparameters for our method were set as \( \beta = 0.4 \), \( \gamma = 0.5 \), and \( p = 0.5 \).}
\label{tab:cleanaccuracy}
\resizebox{\columnwidth}{!}{
\begin{tabular}{llcccccc}
\toprule
Dataset      & $\alpha$ & nwFedAvg  & FedAvg  & FedProx  & FedNova  & SCAFFOLD & Ours \\
\midrule
CIFAR-10     & \textit{0.125} & 51.6  & 52.44  & 52.13  & 52.14  & 52.11  & \textbf{60.23} \\
             & \textit{0.3}   & 56.9  & 59.79  & 57.99  & 58.21  & 57.93  & \textbf{66.36} \\
             & \textit{0.5}   & 60.6  & 61.44  & 60.78  & 60.69  & 61.03  & \textbf{65.5} \\
\midrule
CIFAR-100    & \textit{0.125} & 27.63 & 26.48  & 26.91  & 27.51  & 26.75  & \textbf{29.03} \\
             & \textit{0.3}   & 28.78 & 28.91  & 29.17  & 28.0   & 29.04  & \textbf{35.52} \\
             & \textit{0.5}   & 29.62 & 30.0   & 30.05  & 29.75  & 29.94  & \textbf{36.52} \\
\midrule
FashionMNIST & \textit{0.125} & \textbf{80.43} & 79.51  & 79.14  & 80.2   & 77.97  & 78.21 \\
             & \textit{0.3}   & 87.83 & 88.14  & 87.65  & 87.41  & 87.35  & \textbf{89.32} \\ 
             & \textit{0.5}   & 88.5  & 88.43  & 87.67  & 88.73  & 88.6   & \textbf{89.72} \\
\bottomrule
\end{tabular}
}
\end{table}

\begin{table}[h!]
\centering
\caption{Percentage improvement in accuracy of our method compared to other aggregation methods for CIFAR-10, CIFAR-100, and FashionMNIST across different $\alpha$ values. A positive percentage indicates that our method outperforms the respective baseline.}
\label{tab:accuracy_improvement}
\resizebox{\columnwidth}{!}{
\begin{tabular}{llccccc}
\toprule
Dataset      & $\alpha$ & nwFedAvg & FedAvg & FedProx & FedNova & SCAFFOLD \\
\midrule
CIFAR-10     & \textit{0.125} & \%16.72 & \%14.85 & \%15.53 & \%15.51 & \%15.58 \\
             & \textit{0.3}   & \%16.61 & \%11.00 & \%14.44 & \%13.98 & \%14.54 \\
             & \textit{0.5}   & \%8.08  & \%6.60  & \%7.78  & \%7.91  & \%7.32  \\
\midrule
CIFAR-100    & \textit{0.125} & \%5.07  & \%9.63  & \%7.88  & \%5.53  & \%8.51  \\
             & \textit{0.3}   & \%23.41 & \%22.87 & \%21.78 & \%26.86 & \%22.31 \\
             & \textit{0.5}   & \%23.29 & \%21.73 & \%21.55 & \%22.30 & \%21.94 \\
\midrule
FashionMNIST & \textit{0.125} & \%-2.76 & \%-1.63 & \%-1.18 & \%-2.48 & \%0.31  \\
             & \textit{0.3}   & \%1.70  & \%1.34  & \%1.90  & \%2.19  & \%2.25  \\
             & \textit{0.5}   & \%1.38  & \%1.46  & \%2.34  & \%1.12  & \%1.26  \\
\bottomrule
\end{tabular}
}
\end{table}




\begin{figure*}[t]
    \centering
    \begin{subfigure}[b]{0.46\textwidth}
        \centering
        \includegraphics[width=\linewidth]{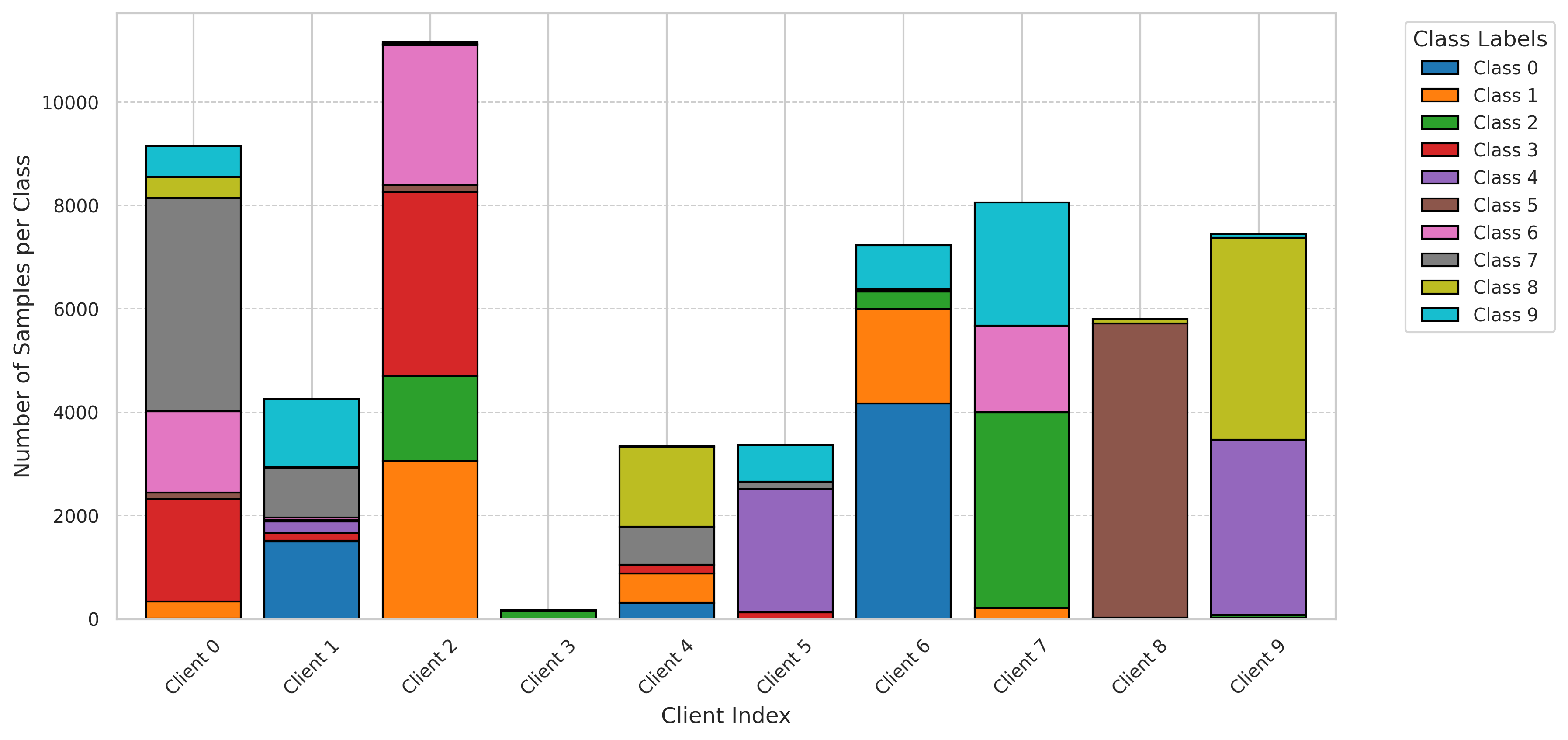}
        \caption{Class distribution across clients in the federated Fashion-MNIST dataset for $\alpha = 0.125$. The strong imbalance in class distribution among clients reflects a highly non-IID setting, which challenges model convergence and generalization.}
        \label{fig:distribution_alpha_0.125}
    \end{subfigure}
    \hfill
    \begin{subfigure}[b]{0.46\textwidth}
        \centering
        \includegraphics[width=\linewidth]{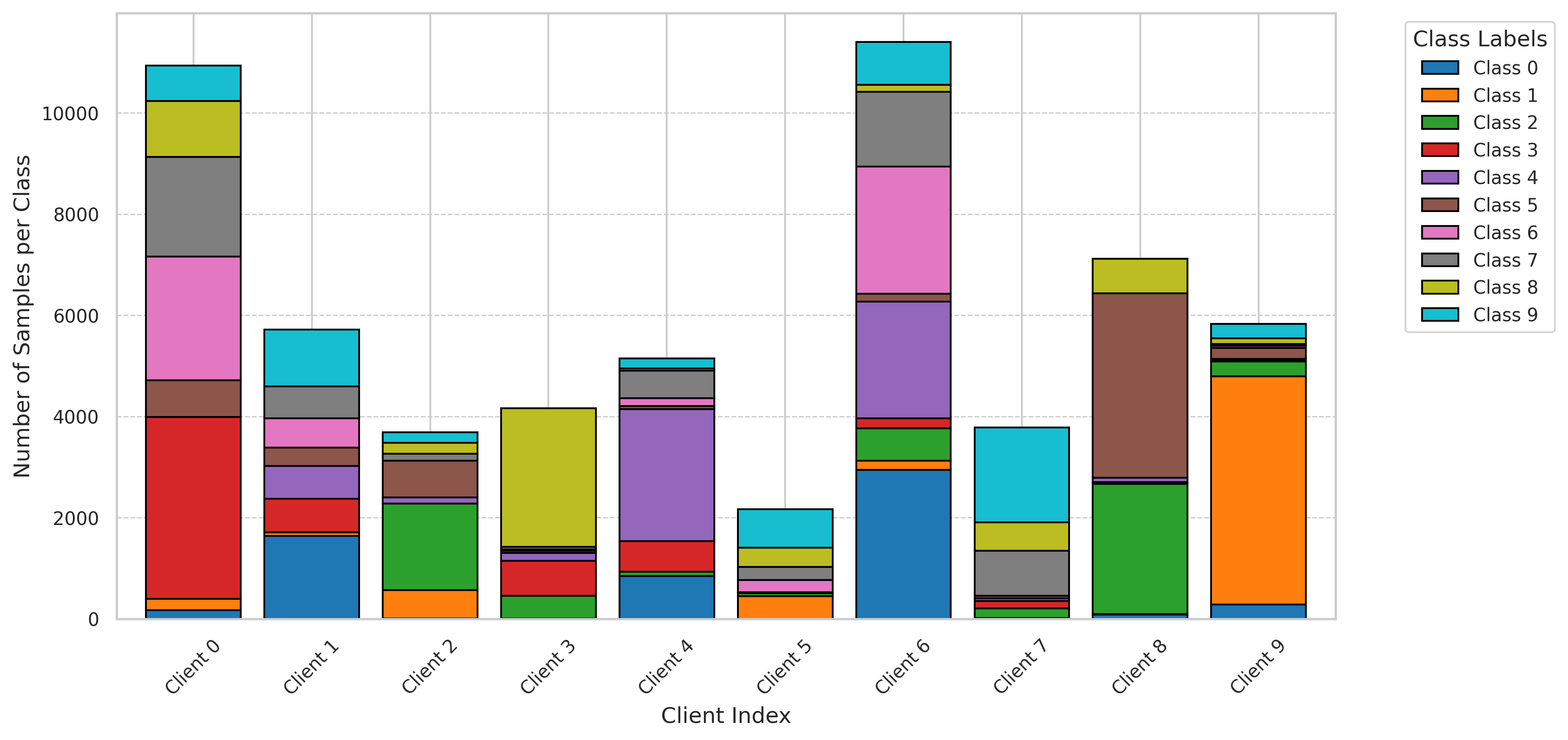}
        \caption{Class distribution across clients in the federated Fashion-MNIST dataset for $\alpha = 0.3$. Compared to $\alpha = 0.125$, the class distribution remains non-uniform but is slightly more balanced, reducing the severity of data heterogeneity.}
        \label{fig:distribution_alpha_0.3}
    \end{subfigure}
    \vskip 0.3cm
    \begin{subfigure}[b]{0.46\textwidth}
        \centering
        \includegraphics[width=\linewidth]{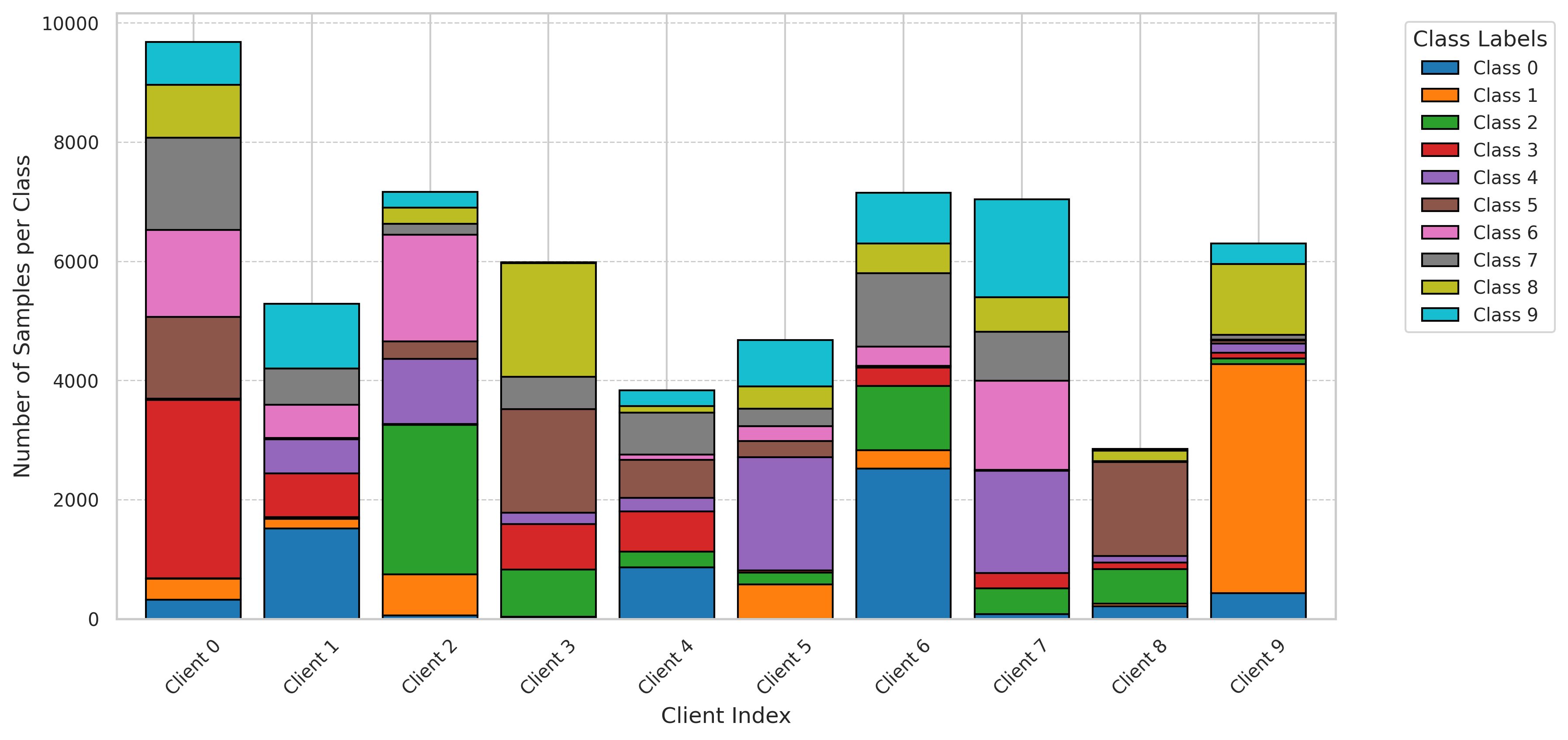}
        \caption{Class distribution across clients in the federated Fashion-MNIST dataset for $\alpha = 0.5$. At this level of $\alpha$, the distribution is closer to an IID setting, mitigating extreme client disparities and potentially improving model convergence.}
        \label{fig:distribution_alpha_0.5}
    \end{subfigure}
    \caption{Comparison of class distributions across clients in the federated Fashion-MNIST dataset for different values of $\alpha$.}
    \label{fig:distribution_alpha}
\end{figure*}

\begin{figure*}[t]
    \centering
    \begin{subfigure}[b]{0.48\textwidth}
        \centering
        \includegraphics[width=\linewidth]{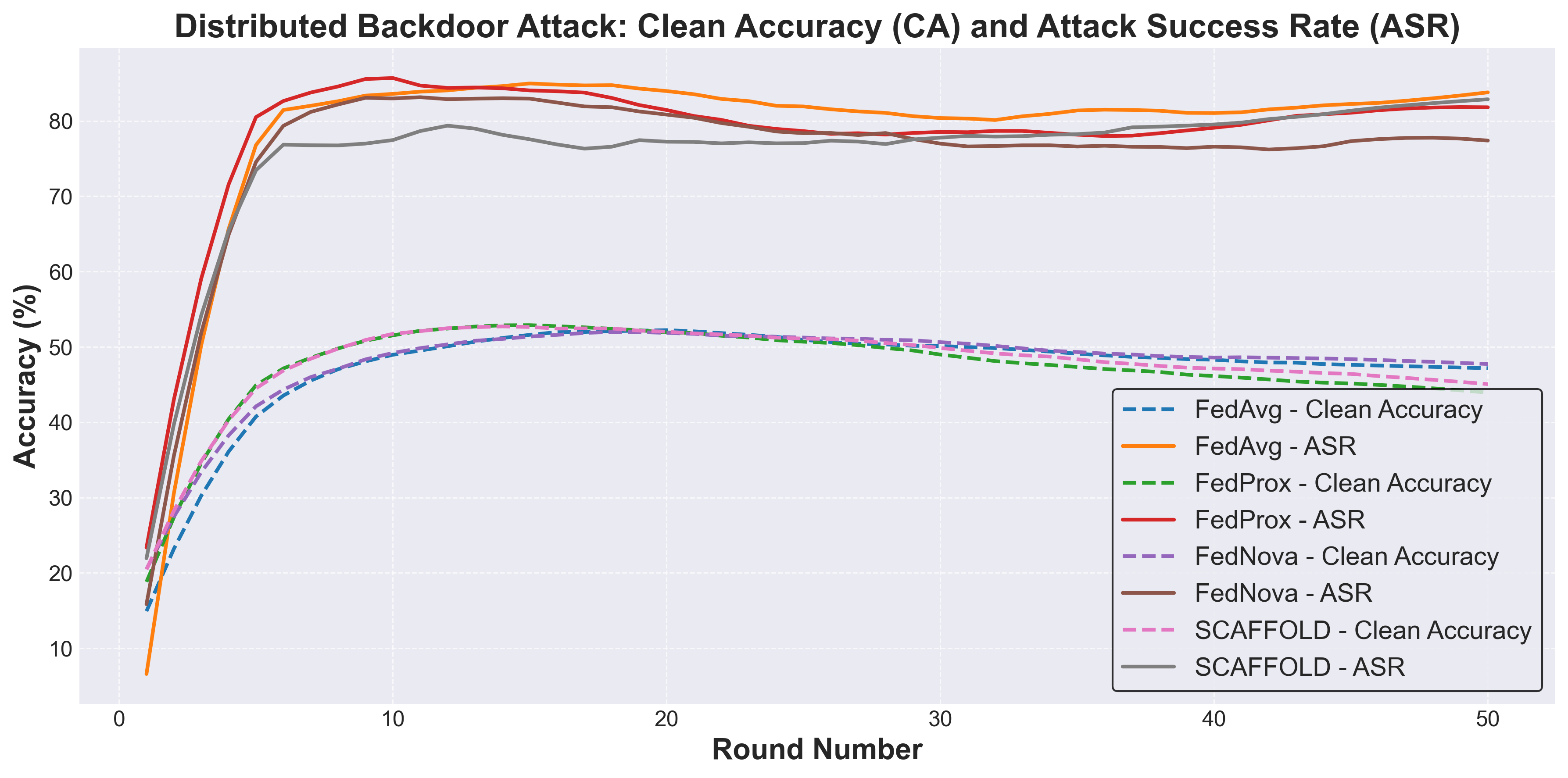}
        \caption{Comparison of baseline federated aggregation methods under a CIFAR-10 distributed backdoor attack scenario. The experiment is conducted with a Dirichlet data distribution ($\alpha = 0.3$), 30\% malicious clients, and a 20\% backdoor injection rate. The figure illustrates the Clean Accuracy (CA) and Attack Success Rate (ASR) across training rounds for five baseline methods: nwFedAvg, FedAvg, FedProx, FedNova, and SCAFFOLD. }
        \label{fig:others_dba}
    \end{subfigure}
    \hfill
    \begin{subfigure}[b]{0.48\textwidth}
        \centering
        \includegraphics[width=\linewidth]{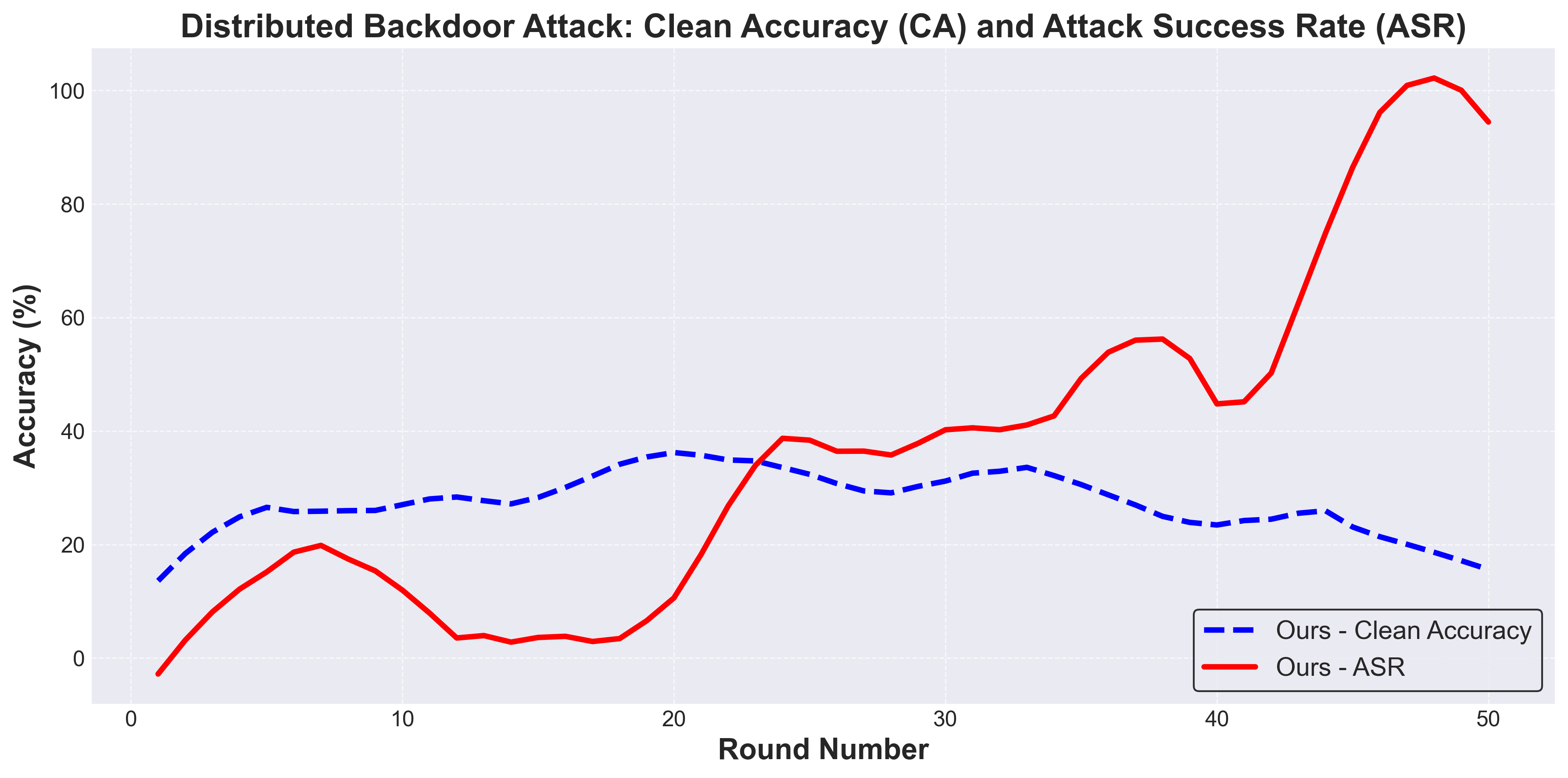}
        \caption{Evaluation of our proposed aggregation method under a CIFAR-10 distributed backdoor attack scenario. The experiment follows the same settings as Figure~\ref{fig:others_dba}, with a Dirichlet data distribution ($\alpha = 0.3$), 30\% malicious clients, and a 20\% backdoor injection rate. The figure compares Clean Accuracy (CA) and Attack Success Rate (ASR) over training rounds.}
        \label{fig:ours_dba}
    \end{subfigure}
\end{figure*}

\begin{table*}[h!!]
\caption{Accuracy (A) and Attack Success Rate (ASR) under Distributed Backdoor Attack for CIFAR-10, CIFAR-100, and FashionMNIST with $\alpha = \textit{0.125}$, $\alpha = \textit{0.3}$, and $\alpha = \textit{0.5}$, client number=\textit{10}, across different aggregation methods. The hyperparameters for our method were set as \( \beta = 0.4 \), \( \gamma = 0.5 \), and \( p = 0.5 \). The malicious client ratio was set to \textit{30\%}, and the backdoored data ratio within these clients was \textit{20\%}.}
\label{tab:backdoor_attack_03} 
\centering
\resizebox{0.74\textwidth}{!}{ 
\begin{tabular}{llcccccccccccc}
\toprule
Dataset      & $\alpha$ & \multicolumn{2}{c}{nwFedAvg} & \multicolumn{2}{c}{FedAvg} & \multicolumn{2}{c}{FedProx} & \multicolumn{2}{c}{FedNova} & \multicolumn{2}{c}{SCAFFOLD} & \multicolumn{2}{c}{Ours} \\
\cmidrule(lr){3-4} \cmidrule(lr){5-6} \cmidrule(lr){7-8} \cmidrule(lr){9-10} \cmidrule(lr){11-12} \cmidrule(lr){13-14}
             &          & A (\%) & ASR (\%) & A (\%) & ASR (\%) & A (\%) & ASR (\%) & A (\%) & ASR (\%) & A (\%) & ASR (\%) & A (\%) & ASR (\%) \\
\midrule
CIFAR-10     & \textit{0.125}   & 48.16    & 82.87    & 49.28    & 84.79    & 49.83    & 84.09    & 48.87    & 87.11    & 50.05    & 80.55   & 11.67 & 99.98 \\   
             & \textit{0.3}      & 54.50    & 83.96    & 55.05    & 77.29    & 55.18   & 84.18    & 53.79    & 85.98    & 54.24    & 82.06    & 16.98 & 98.14 \\      
             & \textit{0.5}      & 56.72    & 84.48    & 57.21    & 87.01    & 56.60   & 87.38    & 57.20    & 83.08    & 56.35    & 83.14   & 32.28 & 89.75 \\
\midrule
CIFAR-100    & \textit{0.125}    & 20.96    & 60.05    & 21.53    & 49.94    & 22.21   & 53.95    & 21.06    & 56.06    & 22.16    & 44.05   & 1.02 & 99.99 \\   
             & \textit{0.3}      & 23.51    & 55.10    & 25.10    & 42.00    & 24.87   & 49.64    & 24.03    & 51.51    & 29.20    & 72.18   & 1.18 & 99.91 \\   
             & \textit{0.5}      & 30.56    & 78.02    & 29.78    & 75.43    & 29.53   & 75.46    & 30.21    & 76.23    & 29.80    & 76.27   & 1.04 & 99.95 \\   
\midrule
FashionMNIST & \textit{0.125}    & 71.86    & 97.77    & 72.84    & 91.73    & 72.07   & 90.27    & 72.38    & 97.37    & 74.74    & 95.62   & 10.34 & 100 \\
             & \textit{0.3}      & 85.26    & 99.06    & 85.40    & 99.53    & 85.35   & 99.95    & 84.08    & 99.22    & 84.54    & 99.65   & 20.68 & 100 \\
             & \textit{0.5}      & 86.27    & 98.60    & 86.57    & 98.00    & 86.42   & 98.25    & 86.62    & 96.64    & 86.21    & 97.98   & 44.81 & 99.94 \\
\bottomrule
\end{tabular}}

\end{table*}

\begin{table*}[h!]
\caption{Accuracy (A) and Attack Success Rate (ASR) under Distributed Backdoor Attack for CIFAR-10 and FashionMNIST with $\alpha = \textit{0.125}$, $\alpha = \textit{0.3}$, and $\alpha = \textit{0.5}$, client number=\textit{10}, across different aggregation methods. The hyperparameters for our method were set as \( \beta = 0.4 \), \( \gamma = 0.5 \), and \( p = 0.5 \). The malicious client ratio was set to = 0.1 and backdoor data ratio = 0.2.}
\label{tab:backdoor_attack_mal_01}
\centering
\resizebox{0.74\textwidth}{!}{ 
\begin{tabular}{llcccccccccccc}
\toprule
Dataset      & $\alpha$ & \multicolumn{2}{c}{nwFedAvg} & \multicolumn{2}{c}{FedAvg} & \multicolumn{2}{c}{FedProx} & \multicolumn{2}{c}{FedNova} & \multicolumn{2}{c}{SCAFFOLD} & \multicolumn{2}{c}{Ours} \\
\cmidrule(lr){3-4} \cmidrule(lr){5-6} \cmidrule(lr){7-8} \cmidrule(lr){9-10} \cmidrule(lr){11-12} \cmidrule(lr){13-14}
             &          & A (\%) & ASR (\%) & A (\%) & ASR (\%) & A (\%) & ASR (\%) & A (\%) & ASR (\%) & A (\%) & ASR (\%) & A (\%) & ASR (\%) \\
\midrule
CIFAR-10     & \textit{0.125}   & 47.79  & 40.67  & 47.6  & 36.73  & 48.64  & 34.66  & 48.04  & 35.71 & 48.2  & 38.87  & 35.17 & 12.63 \\
             & \textit{0.3}      & 55.35  & 40.63  & 53.78  & 36.93  & 55.08  & 37.63  &  55.16  & 35.45  & 53.91  & 34.74  & 53.48 & 7.77 \\
             & \textit{0.5}      & 57.95  & 35.72  & 56.45  & 36.98  & 55.62  & 33.35  & 57.14  & 32.69  & 56.8  &  32.28  & 59.62 & 11.27 \\
\midrule
FashionMNIST & \textit{0.125}    & 78.99  & 64.6  & 76.35  & 37.15  & 77.29  & 33.85  & 78.75  & 55.3  &  77.12  & 39.21  & 44.55 & 4.54 \\
             & \textit{0.3}      & 86.68  & 84.56 & 87.03 & 71.84  & 86.8  & 81.47  & 87.34  & 78.79  & 87.21 & 68.89  & 80.82 & 10.5 \\
             & \textit{0.5}      &  87.89  & 72.35 & 87.35 & 57.68  & 87.46  & 59.43  &  87.76  & 70.94  &  87.28  & 47.7  & 81.07 & 11.71 \\
\bottomrule
\end{tabular}}
\end{table*}

\subsubsection{\textbf{Distributed Backdoor Attacks}}
\label{sec:dba_results}
We evaluate the robustness of our aggregation strategy under a distributed backdoor attack and compare it with traditional aggregation methods, including \textit{nwFedAvg}, \textit{FedAvg}, \textit{FedProx}, \textit{FedNova} and \textit{SCAFFOLD}. The attack is conducted under a controlled setting where 30\% of the clients are malicious, and 20\% of their training data contains backdoor triggers. Our objective is twofold: (i) to examine whether traditional aggregation methods can resist the backdoor attack and (ii) to analyze how our proposed aggregation method inherently responds to the same attack under different non-IID settings (\( \alpha = 0.125, 0.3, 0.5 \)). Additionally, Table~\ref{tab:accuracy_diff} presents the \textit{absolute accuracy degradation} between the clean and attacked models. Furthermore, Figure~\ref{fig:others_dba} and Figure~\ref{fig:distribution_alpha_0.3} illustrate how ASR and A evolve over time.

The results in Table~\ref{tab:backdoor_attack_03} show that traditional aggregation methods exhibit consistently high ASR values across all datasets, confirming that the backdoor attack is highly effective in these settings. Specifically, for CIFAR-10, ASR remains between 77.29\% and 87.38\%, while for CIFAR-100, ASR ranges from 42.00\% to 78.02\%. In FashionMNIST, the attack is even more successful, achieving over 90\% ASR in all cases and reaching 99.95\% in some configurations.

Furthermore, as shown in Table~\ref{tab:accuracy_diff}, the absolute accuracy difference between the clean and attacked models remains minimal for traditional aggregation methods, typically below 5\% in most cases. This indicates that the attack is successful, as it maintains high classification accuracy on clean data while achieving high ASR, effectively embedding the backdoor without disrupting normal model performance. The evolution of ASR and clean accuracy over time, depicted in Figure~\ref{fig:others_dba}, reveals that in traditional methods, ASR increases as clean accuracy improves. Once the model reaches a learning plateau, both ASR and A stabilize, indicating that the attack remains persistent and undetected.

When applying the same attack to our aggregation strategy, we observe a fundamentally different pattern in both ASR evolution and model performance. Unlike traditional aggregation methods, which allow adversaries to embed backdoors with minimal accuracy degradation, our method exhibits a distinctive learning behavior that disrupts the attack's success.

As shown in Figure~\ref{fig:distribution_alpha_0.3}, in our method, ASR does not increase significantly in the early training stages, even as the model begins learning. Instead, ASR remains low until the model reaches a learning plateau. At this point, the attack enters a critical phase: ASR increases, while clean accuracy begins to sharply decline. This distinct behavior indicates that the backdoor injection is not successful in a traditional sense; rather, it destabilizes the entire learning process, making the attack detectable and ultimately ineffective. Moreover, as shown in Table~\ref{tab:accuracy_diff}, the absolute accuracy drop in our method is substantially larger compared to traditional approaches. For CIFAR-10, the accuracy drops by -48.56 to -49.38, making it impractical for an adversary to maintain both a high ASR and model utility. Similarly, in CIFAR-100, our method enforces a accuracy degradation of -28.01 to -35.48, while in FashionMNIST, the model undergoes a drastic accuracy drop of up to -68.64. Following this observation, we noted that the distributed backdoor attack setting (30\% malicious clients, 20\% backdoored data) led to a significant degradation in clean accuracy for our method, effectively making the attack detectable and ultimately ineffective. To further investigate the resilience of our aggregation strategy against more subtle attack, we conducted an additional experiment with a lower malicious client ratio (10\%) while maintaining the same backdoor data ratio (20\%). This adjustment aimed to assess whether the attack could remain effective while preventing a drastic clean accuracy drop, which would make it more difficult to detect. Table~\ref{tab:backdoor_attack_mal_01} presents the results of this modified attack setting for CIFAR-10 and FashionMNIST under different non-IID levels (\( \alpha = 0.125, 0.3, 0.5 \)), comparing our method with traditional aggregation methods. Notably, CIFAR-100 is not included in this table because, under this lower-intensity attack, the backdoor injection was ineffective across all aggregation methods, making it uninformative for analysis. The results show that, for CIFAR-10 and FashionMNIST, traditional aggregation methods still maintain high ASR while preserving clean accuracy. Specifically, in CIFAR-10, ASR remains between 32.28\% and 40.67\%, and in FashionMNIST, it ranges from 33.85\% to 84.56\%, depending on the aggregation method. These results indicate that the attack is still successful in traditional methods, allowing adversaries to introduce backdoor behavior without drastically affecting classification performance. For our method, the results in Table~\ref{tab:backdoor_attack_mal_01} demonstrate a drastically lower ASR compared to traditional approaches, with values dropping to below 12.63\% in CIFAR-10 and below 11.71\% in FashionMNIST across all non-IID settings. This is a major contrast to the high ASR observed in traditional aggregation methods, reinforcing that our approach continues to resist the backdoor attack even under a milder adversarial setting.

\subsubsection{\textbf{Understanding Accuracy Degradation in DBA: Why Does Accuracy Drop While ASR Remains High in Table~\ref{tab:backdoor_attack_03}?}}\label{subsec:backdoor_attack_03}
This behavior can be attributed to the selective nature of our gradient-based masking strategy. During the initial training stages, the most significant gradients are retained and aggregated, effectively isolating backdoor updates from clean ones. This can easily be observed in \ref{fig:ours_dba}. As a result, the model remains unpoisoned while actively learning. However, as training progresses and the model reaches a convergence plateau, the overall gradient magnitudes decrease, making backdoor-related gradients more prominent relative to their clean counterparts. Once these gradients surpass the zip percent threshold, they are incorporated into the mask. Due to the persistence of masked updates across rounds, backdoor gradients accumulate over time, ultimately disrupting clean accuracy. This issue can be mitigated through early stopping mechanisms, adaptive accuracy-based anomaly detection, or by identifying and filtering participants whose updates contribute disproportionately to accuracy degradation.


\subsubsection{\textbf{Convergence Prevention Attacks}}
\label{sec:convergence_prevention}
To evaluate the impact of convergence prevention attacks on different aggregation strategies, we measure the accuracy degradation compared to the \textit{FedAvg} baseline. The results for CIFAR-10, CIFAR-100, and FashionMNIST across varying non-IID settings (\( \alpha = 0.125, 0.3, 0.5 \)) are presented in Table~\ref{tab:cp_attack_diff}. Higher accuracy degradation indicates a greater vulnerability to convergence disruption, while lower degradation suggests increased resilience. The results in Table~\ref{tab:cp_attack_diff} show that our method consistently demonstrates higher resilience to convergence prevention attacks compared to other aggregation strategies. This is evident from the relatively lower accuracy degradation across all datasets and non-IID levels.  For CIFAR-10, our method exhibits the least performance drop across all three \(\alpha\) values, with a significant margin of improvement over other methods, particularly at \(\alpha = 0.5\), where it outperforms the next-best approach FedNova by over 12.5 percentage points  (\%15.18 vs. \%2.66). This indicates that our approach is more robust in mitigating the effects of malicious updates in moderately heterogeneous settings. For CIFAR-100, our method again achieves the smallest performance degradation, with particularly strong resilience at higher non-IID levels (\(\alpha = 0.3, 0.5\)). Notably, other methods such as SCAFFOLD and FedNova struggle in this setting, often exhibiting negative accuracy differences. In the case of FashionMNIST, while all methods show greater vulnerability at \(\alpha = 0.125\), our method exhibits more stable performance at higher \(\alpha\) values, maintaining accuracy improvements even under attack conditions. FedProx and SCAFFOLD display greater instability, as reflected in their highly negative accuracy differences at lower \(\alpha\). These results proves that aggregation methods relying on explicit metadata exchange (such as FedNova and SCAFFOLD) struggle more against convergence prevention attacks, likely due to their dependency on dataset sizes, local step counts, or control variates—making them more susceptible to manipulation. Conversely, our method, which does not rely on these additional parameters, is able to retain greater stability under adversarial conditions, reinforcing its effectiveness against convergence prevention attacks.  

It is important to note that mitigating data poisoning attacks in non-IID federated learning remains a difficult problem \cite{247652, karimireddy2023byzantinerobustlearningheterogeneousdatasets, ghosh2019}. Our method is not specifically designed as a defense mechanism against such attacks. Instead, through these experiments, we evaluate its behavior in comparison to other aggregation strategies.

Our experimental results demonstrate that our method outperforms baseline approaches in both standard (non-attacked) scenarios and robustness against adversarial attacks. In subtle backdoor attack scenarios, where traditional methods fail to prevent the attack, our approach effectively suppresses the impact of backdoored updates while maintaining high clean accuracy. As the malicious client ratio increases, the clean model’s accuracy declines, ultimately making the backdoor attack detectable and ineffective. In Section-\ref{subsec:backdoor_attack_03}, we reason this behavior and propose a mitigation strategy to address it. For convergence prevention attacks, our method significantly reduces the attack’s effectiveness compared to traditional approaches. This is primarily due to its gradient masking mechanism, which prioritizes class-relevant updates, thereby minimizing the influence of adversarial gradients. As a result, our method not only mitigates the attack’s impact but also preserves clean model accuracy better than other methods, reinforcing its overall effectiveness in federated learning settings.

\begin{table}[h!]
\centering
\caption{Negative Absolute Accuracy Differences between Clean and Distributed Backdoor Attack for CIFAR-10, CIFAR-100, and FashionMNIST across different aggregation methods. A higher decrease in accuracy indicates a failure of the backdoor attack.}
\label{tab:accuracy_diff}
\resizebox{\columnwidth}{!}{
\begin{tabular}{llcccccc}
\toprule
Dataset      & $\alpha$ & nwFedAvg  & FedAvg  & FedProx  & FedNova  & SCAFFOLD  & Ours \\
\midrule
CIFAR-10     & \textit{0.125} & -3.44  & -3.16  & -2.30  & -3.27  & -2.06  & -48.56  \\
             & \textit{0.3}   & -2.40  & -4.74  & -2.81  & -4.42  & -3.69  & -49.38  \\
             & \textit{0.5}   & -3.88  & -4.23  & -4.18  & -3.49  & -4.68  & -33.22  \\
\midrule
CIFAR-100    & \textit{0.125} & -6.67  & -4.95  & -4.70  & -6.45  & -4.59  & -28.01  \\
             & \textit{0.3}   & -5.27  & -3.81  & -4.30  & -3.97  & -0.16  & -34.34  \\
             & \textit{0.5}   & -0.94  & -0.22  & -0.52  & -0.46  & -0.14  & -35.48  \\
\midrule
FashionMNIST & \textit{0.125} & -8.57  & -7.63  & -6.07  & -7.82  & -3.23  & -67.87  \\
             & \textit{0.3}   & -2.57  & -1.87  & -2.30  & -3.33  & -2.81  & -68.64  \\
             & \textit{0.5}   & -2.23  & -1.86  & -1.25  & -2.09  & -2.39  & -44.91  \\
\bottomrule
\end{tabular}
}
\end{table}


\begin{table}[h!]
\centering
\caption{Final Accuracy Differences Compared to FedAvg Baseline under Convergence Prevention Attack for CIFAR-10, CIFAR-100, and FashionMNIST.}
\label{tab:cp_attack_diff}
\resizebox{\columnwidth}{!}{
\begin{tabular}{llccccc}
\toprule
Dataset      & $\alpha$ & nwFedAvg  & FedProx  & FedNova  & SCAFFOLD  & Ours \\
\midrule
CIFAR-10     & \textit{0.125} & \%+1.81  & \%-0.58  & \%+0.20  & \%-6.07  & \%+4.43  \\
             & \textit{0.3}   & \%+4.06  & \%+0.12  & \%+1.64  & \%+6.10  & \%+8.00  \\
             & \textit{0.5}   & \%+4.98  & \%+7.25  & \%+2.66  & \%+4.25  & \%+15.18  \\
\midrule
CIFAR-100    & \textit{0.125} & \%+1.67  & \%+0.81  & \%-0.09  & \%-0.73  & \%+7.70  \\
             & \textit{0.3}   & \%+2.07  & \%+1.77  & \%+1.29  & \%+1.04  & \%+8.11  \\
             & \textit{0.5}   & \%-1.21  & \%-0.47  & \%-1.00  & \%-1.62  & \%+6.06  \\
\midrule
FashionMNIST & \textit{0.125} & \%-8.00  & \%-12.24  & \%-1.07  & \%-2.96  & \%-2.95  \\
             & \textit{0.3}   & \%+1.65  & \%+0.91  & \%+0.67  & \%+1.60  & \%+4.05  \\
             & \textit{0.5}   & \%-0.01  & \%+0.04  & \%-0.14  & \%-0.01  & \%+3.95  \\
\bottomrule
\end{tabular}
}
\end{table}

\begin{figure}[t]
    \centering
    \includegraphics[width=0.8\linewidth]{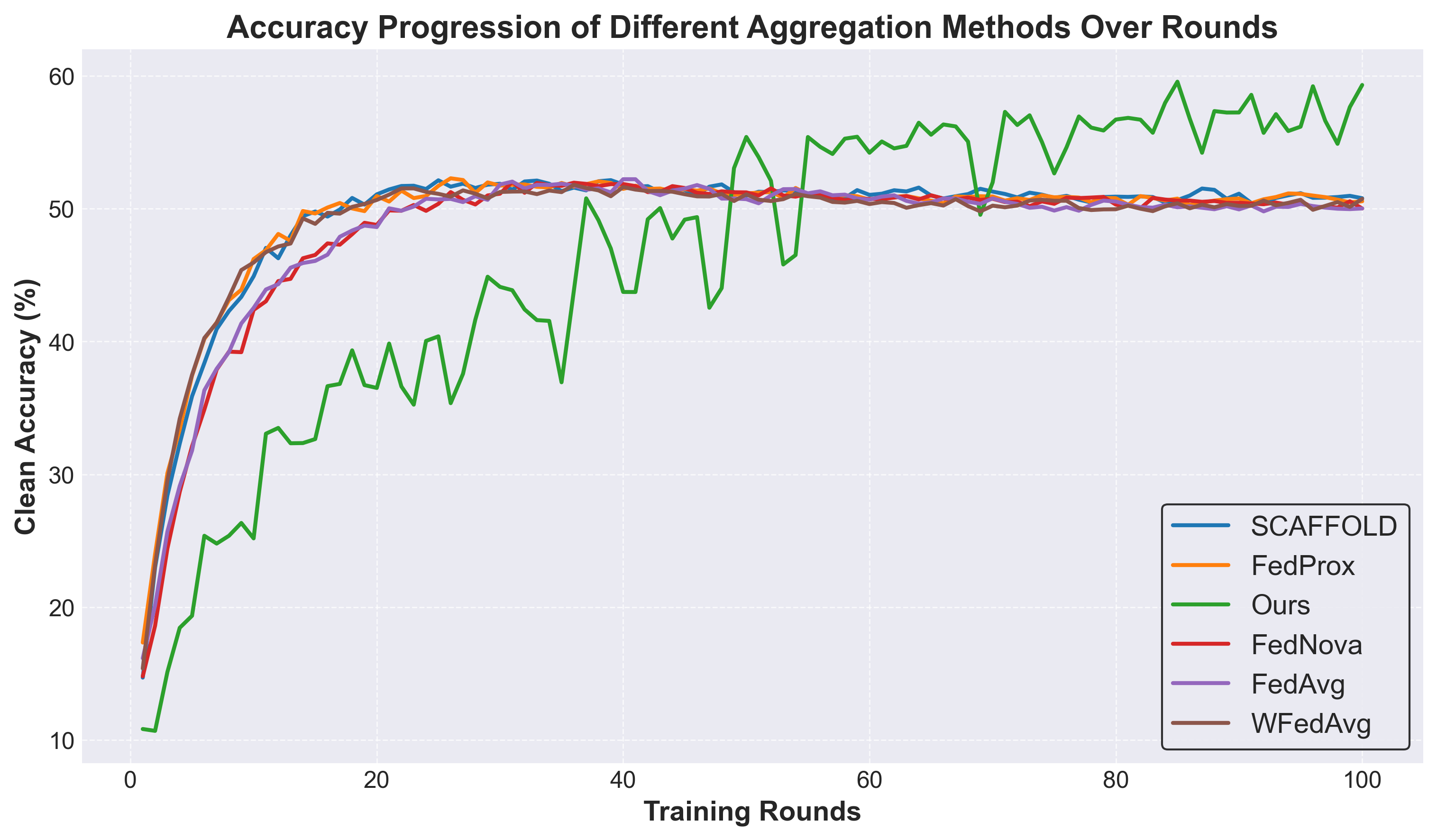}
    \caption{Accuracy progression over training rounds for different aggregation methods on CIFAR-10. The experiment follows a Dirichlet distribution with $\alpha = 0.3$, using 10 clients}
    \label{fig:convergence_plot_for_aggr_methods}
\end{figure}

\subsection{Scalability and Limitations}  
In this section, we analyze the scalability of our approach and discuss its limitations.

\textbf{Computational Overhead and Aggregation Time:}  
Our proposed method introduces additional computational overhead compared to traditional aggregation strategies during aggregation. The primary source of this overhead stems from evaluating locally trained client models against the validation dataset to determine class-specific performance and generate masks. Experimental results conducted on an RTX 4090 GPU show that, in the worst-case scenario, the aggregation time can increase by approximately 43× (see Table~\ref{tab:aggregation_time}) when using our method compared to FedAvg. This increase occurs because every client model is evaluated on every data point. However, this represents an upper bound, as real-world federated learning systems often do not aggregate updates from all clients at each communication round. Instead, a subset of clients is randomly sampled in each round, significantly reducing computational costs. Incorporating such sampling strategies ensures that the method remains scalable for large-scale deployments without compromising model performance.

\begin{table}[h!]
\centering
\caption{Average Aggregation Time on the Server Side for FedAvg and Our Method.}
\label{tab:aggregation_time}
\begin{tabular}{lc}
\toprule
Method      & Aggregation Time (seconds) \\
\midrule
FedAvg      & 0.151 \\
Ours        & 6.450 \\
\bottomrule
\end{tabular}
\end{table}

\textbf{Convergence Time:} We observed that our method generally exhibits a longer convergence time compared to baseline approaches (See Fig-\ref{fig:convergence_plot_for_aggr_methods}) . This slower convergence can be attributed to the \textit{selective parameter updates}. The masking process prunes and scales parameters, reducing the volume of updates applied at each round. While this enhances robustness and noise reduction, it also slows the adaptation rate, particularly in the early rounds. Despite the slower initial convergence, our method consistently achieves higher final accuracy in non-IID scenarios, as shown in Table~\ref{tab:cleanaccuracy}. This suggests that the trade-off between convergence speed and final performance is acceptable, particularly in federated learning settings where the data is highly heterogeneous. Future work may focus on optimizing mask computation, leveraging approximate gradient evaluations, and integrating adaptive masking mechanisms to accelerate convergence while retaining performance benefits.  

\section{Related Work}

Federated Learning was introduced by researchers at Google~\cite{fl-mcmahan}, who proposed the HFL paradigm. In this approach, all clients utilize data from the same feature space, though the data may originate from different sample identities.
In an ideal scenario, the data across clients would be independent and identically distributed (IID). However, this is rarely realistic, as real-world applications commonly involve non-IID data distributions.
The non-IID setting typically presents greater challenges in achieving optimal performance for the final global model when compared to the IID distribution~\cite{zhao2018federated}.
To address this limitation, several advanced solutions have been proposed, including FedProx~\cite{FedProx}, FedNova~\cite{FedNova}, and SCAFFOLD~\cite{scaffold}. These methods aim to enhance the classical FedAvg aggregation strategy by incorporating techniques that account for the high heterogeneity of data across clients.

FedProx introduces a proximal term in the optimization process to address variability in local updates, ensuring more stable convergence even in non-i.i.d. settings. FedNova tackles the issue of client drift by normalizing local updates, aligning them across clients with varying numbers of local epochs. SCAFFOLD, on the other hand, uses control variates to explicitly reduce variance in local updates, mitigating the divergence caused by heterogeneous client data. Collectively, these strategies leverage normalization, variance reduction, or other adjustments to ensure more robust and efficient federated learning, even in highly heterogeneous environments.

All these solutions primarily focus on improving performance without adequately addressing the privacy risks associated with sharing information about clients' data distributions. Such information sharing can expose vulnerabilities to leakage attacks~\cite{arazzi2023blindsage,li2024subject}, potentially compromising the confidentiality of sensitive client data. Moreover, these methods fail to account for the presence of malicious clients that may exploit these weaknesses to perform attacks, such as injecting backdoors into the global model or executing drifting attacks to destabilize training and compromise outcomes.

Backdoor attacks in deep learning were initially introduced by Gu et al.~\cite{badnets}. These attacks involve modifying a subset of training data to implant a hidden functionality within the model, which is triggered during inference. This approach has gained significant attention and has been adapted to federated learning (FL)~\cite{bagdasaryan2020backdoor,wang2020attack,xie2019dba,xu2022more,arazzi2023turning,arazzi2024let}.  In~\cite{bagdasaryan2020backdoor}, the authors proposed a model replacement attack, where adversaries amplify their gradient updates to replace the global model with a compromised version containing a backdoor. Wang et al.~\cite{wang2020attack} developed a theoretical framework showing that models vulnerable to evasion attacks in FL are also susceptible to backdoor attacks. Xie et al.~\cite{xie2019dba} introduced the first distributed backdoor attack in an FL setting by splitting the trigger across multiple malicious clients. This method enables the backdoor to be activated either by individual local triggers or a combined global trigger. Xu et al.~\cite{xu2022more} extended this distributed approach to backdoor federated graph neural networks, demonstrating its versatility in complex FL scenarios.  
On the other side, drifting attacks, such as model poisoning, pose significant threats to the integrity of machine learning systems. In these attacks, malicious participants intentionally introduce harmful updates to the training process, aiming to degrade the model’s overall performance or compromise its reliability. Such attacks can manifest in various ways, including corrupting specific training data or manipulating gradients to introduce biases, rendering the model ineffective or untrustworthy for its intended applications \cite{bhagoji2019analyzing}.
In this paper, we propose an aggregation strategy for non-IID HFL that accounts for the unique challenges of federated settings and actively mitigates potential risks associated with such environments.

\section{Conclusions}
In this work, we presented a novel aggregation method for cross-silo federated learning in non-IID environments. Our experimental results demonstrate that our approach consistently outperforms traditional aggregation strategies by a \textit{significant margin}. 

Furthermore, we showed that our method effectively mitigates backdoor attacks and remains robust against convergence prevention attacks while preserving client privacy. Notably, our approach achieves these improvements solely by analyzing gradients and applying an adaptive masking mechanism, eliminating the need for any client metadata and ensuring privacy remains uncompromised. These findings highlight the potential of our method for enabling private, secure, and scalable federated learning in real-world deployments.

\bibliographystyle{plain}

\begin{thebibliography}{10}
	\providecommand{\url}[1]{\texttt{#1}}
	\providecommand{\urlprefix}{URL }
	\providecommand{\doi}[1]{https://doi.org/#1}
	
	\bibitem{gdpr}
	Regulation (eu) 2016/679 of the european parliament and of the council of 27 april 2016 on the protection of natural persons with regard to the processing of personal data and on the free movement of such data, and repealing directive 95/46/ec (general data protection regulation) (text with eea relevance) (May 2016)
	
	\bibitem{arazzi2023blindsage}
	Arazzi, M., Conti, M., Koffas, S., Krcek, M., Nocera, A., Picek, S., Xu, J.: Blindsage: Label inference attacks against node-level vertical federated graph neural networks. arXiv preprint arXiv:2308.02465  (2023)
	
	\bibitem{arazzi2023turning}
	Arazzi, M., Conti, M., Nocera, A., Picek, S.: Turning privacy-preserving mechanisms against federated learning. In: Proceedings of the 2023 ACM SIGSAC Conference on Computer and Communications Security. pp. 1482--1495 (2023)
	
	\bibitem{arazzi2024let}
	Arazzi, M., Koffas, S., Nocera, A., Picek, S.: Let's focus: Focused backdoor attack against federated transfer learning. arXiv preprint arXiv:2404.19420  (2024)
	
	\bibitem{marco_arazzi_evading_model_poisoning}
	Arazzi, M., Lax, G., Nocera, A.: Evading model poisoning attacks in federated learning by a long-short-term-memory-based approach. Integrated Computer-Aided Engineering  \textbf{0}(0),  10692509241301588 (0). \doi{10.1177/10692509241301588}, \url{https://doi.org/10.1177/10692509241301588}
	
	\bibitem{bagdasaryan2020backdoor}
	Bagdasaryan, E., Veit, A., Hua, Y., Estrin, D., Shmatikov, V.: How to backdoor federated learning. In: International conference on artificial intelligence and statistics. pp. 2938--2948. PMLR (2020)
	
	\bibitem{bhagoji2019analyzing}
	Bhagoji, A.N., Chakraborty, S., Mittal, P., Calo, S.: Analyzing federated learning through an adversarial lens. In: Proceedings of the 36th International Conference on Machine Learning (ICML). PMLR (2019), \url{https://doi.org/10.48550/arXiv.1811.12470}
	
	\bibitem{bonawitz2019towards}
	Bonawitz, K., Eichner, H., Grieskamp, W., Huba, D., Ingerman, A., Ivanov, V., Kiddon, C., Konecny, J., Mazzocchi, S., McMahan, H.B., et~al.: Towards federated learning at scale: System design. arXiv preprint arXiv:1902.01046  (2019)
	
	\bibitem{bonawitz2017practical}
	Bonawitz, K., Ivanov, V., Kreuter, B., Marcedone, A., McMahan, H.B., Patel, S., Ramage, D., Segal, A., Seth, K.: Practical secure aggregation for privacy preserving machine learning. IACR Cryptology ePrint Archive  \textbf{2017}, ~281 (2017)
	
	\bibitem{dinh2022personalizedfederatedlearningmoreau}
	Dinh, C.T., Tran, N.H., Nguyen, T.D.: Personalized federated learning with moreau envelopes (2022), \url{https://arxiv.org/abs/2006.08848}
	
	\bibitem{247652}
	Fang, M., Cao, X., Jia, J., Gong, N.: Local model poisoning attacks to {Byzantine-Robust} federated learning. In: 29th USENIX Security Symposium (USENIX Security 20). pp. 1605--1622. USENIX Association (Aug 2020), \url{https://www.usenix.org/conference/usenixsecurity20/presentation/fang}
	
	\bibitem{ghosh2019}
	Ghosh, A., Hong, J., Yin, D., Ramchandran, K.: Robust federated learning in a heterogeneous environment (2019), \url{https://arxiv.org/abs/1906.06629}
	
	\bibitem{badnets}
	Gu, T., Dolan-Gavitt, B., Garg, S.: Badnets: Identifying vulnerabilities in the machine learning model supply chain. arXiv preprint arXiv:1708.06733  (2017)
	
	\bibitem{han2024badsflbackdoorattackscaffold}
	Han, X., Zhang, X., Lan, X., Wang, H., Xu, S., Ren, S., Zeng, J., Wu, M., Heinrich, M., Zhang, T.: Badsfl: Backdoor attack against scaffold federated learning (2024), \url{https://arxiv.org/abs/2411.16167}
	
	\bibitem{huang2022crosssilofederatedlearningchallenges}
	Huang, C., Huang, J., Liu, X.: Cross-silo federated learning: Challenges and opportunities (2022), \url{https://arxiv.org/abs/2206.12949}
	
	\bibitem{huang2021personalizedcrosssilofederatedlearning}
	Huang, Y., Chu, L., Zhou, Z., Wang, L., Liu, J., Pei, J., Zhang, Y.: Personalized cross-silo federated learning on non-iid data (2021), \url{https://arxiv.org/abs/2007.03797}
	
	\bibitem{karimireddy2023byzantinerobustlearningheterogeneousdatasets}
	Karimireddy, S.P., He, L., Jaggi, M.: Byzantine-robust learning on heterogeneous datasets via bucketing (2023), \url{https://arxiv.org/abs/2006.09365}
	
	\bibitem{scaffold}
	Karimireddy, S.P., Kale, S., Mohri, M., Reddi, S.J., Stich, S.U., Suresh, A.T.: Scaffold: Stochastic controlled averaging for federated learning. arXiv preprint arXiv:1910.06378  (2020), \url{https://arxiv.org/abs/1910.06378}
	
	\bibitem{konecny2016federated}
	Konecn{\`y}, J., McMahan, H.B., Ramage, D., Richt{\'a}rik, P.: Federated optimization: Distributed machine learning for on-device intelligence. arXiv preprint arXiv:1610.02527  (2016)
	
	\bibitem{Konecn2016FederatedLS}
	Konecn{\'y}, J., McMahan, H.B., Yu, F.X., Richt{\'a}rik, P., Suresh, A.T., Bacon, D.: Federated learning: Strategies for improving communication efficiency. ArXiv  \textbf{abs/1610.05492} (2016), \url{https://api.semanticscholar.org/CorpusID:14999259}
	
	\bibitem{Krizhevsky2009LearningML}
	Krizhevsky, A.: Learning multiple layers of features from tiny images (2009), \url{https://api.semanticscholar.org/CorpusID:18268744}
	
	\bibitem{Li_2022}
	Li, C., Li, G., Varshney, P.K.: Decentralized federated learning via mutual knowledge transfer. IEEE Internet of Things Journal  \textbf{9}(2),  1136–1147 (Jan 2022). \doi{10.1109/jiot.2021.3078543}, \url{http://dx.doi.org/10.1109/JIOT.2021.3078543}
	
	\bibitem{li2024subject}
	Li, J., Arazzi, M., Nocera, A., Conti, M.: Subject data auditing via source inference attack in cross-silo federated learning. arXiv preprint arXiv:2409.19417  (2024)
	
	\bibitem{li2022federated}
	Li, Q., Diao, Y., Chen, Q., He, B.: Federated learning on non-iid data silos: An experimental study. In: 2022 IEEE 38th international conference on data engineering (ICDE). pp. 965--978. IEEE (2022)
	
	\bibitem{li2019federated}
	Li, T., Sahu, A.K., Talwalkar, A., Smith, V.: Federated learning: Challenges, methods, and future directions. arXiv preprint arXiv:1908.07873  (2019)
	
	\bibitem{li2021dittofairrobustfederated}
	Li, T., Hu, S., Beirami, A., Smith, V.: Ditto: Fair and robust federated learning through personalization (2021), \url{https://arxiv.org/abs/2012.04221}
	
	\bibitem{FedProx}
	Li, T., Sahu, A.K., Zaheer, M., Sanjabi, M., Talwalkar, A., Smith, V.: Federated optimization in heterogeneous networks. In: Proceedings of Machine Learning and Systems (MLSys) (2020), \url{https://arxiv.org/abs/1812.06127}
	
	\bibitem{liu2022privacy}
	Liu, K., Hu, S., Wu, S.Z., Smith, V.: On privacy and personalization in cross-silo federated learning. Advances in neural information processing systems  \textbf{35},  5925--5940 (2022)
	
	\bibitem{fl-mcmahan}
	McMahan, B., Moore, E., Ramage, D., Hampson, S., Arcas, B.A.y.: {Communication-Efficient Learning of Deep Networks from Decentralized Data}. In: Singh, A., Zhu, J. (eds.) Proceedings of the 20th International Conference on Artificial Intelligence and Statistics. Proceedings of Machine Learning Research, vol.~54, pp. 1273--1282. PMLR (20--22 Apr 2017), \url{https://proceedings.mlr.press/v54/mcmahan17a.html}
	
	\bibitem{mei2023privacy}
	Mei, H., Li, G., Wu, J., Zheng, L.: Privacy inference-empowered stealthy backdoor attack on federated learning under non-iid scenarios. In: 2023 International Joint Conference on Neural Networks (IJCNN). pp. 1--10. IEEE (2023)
	
	\bibitem{reyes2021precisionweightedfederatedlearning}
	Reyes, J., Jorio, L.D., Low-Kam, C., Kersten-Oertel, M.: Precision-weighted federated learning (2021), \url{https://arxiv.org/abs/2107.09627}
	
	\bibitem{sattler2019clusteredfederatedlearningmodelagnostic}
	Sattler, F., Müller, K.R., Samek, W.: Clustered federated learning: Model-agnostic distributed multi-task optimization under privacy constraints (2019), \url{https://arxiv.org/abs/1910.01991}
	
	\bibitem{Shejwalkar2021ManipulatingTB}
	Shejwalkar, V., Houmansadr, A.: Manipulating the byzantine: Optimizing model poisoning attacks and defenses for federated learning. Proceedings 2021 Network and Distributed System Security Symposium  (2021), \url{https://api.semanticscholar.org/CorpusID:231861235}
	
	\bibitem{stripelis2022semi}
	Stripelis, D., Thompson, P.M., Ambite, J.L.: Semi-synchronous federated learning for energy-efficient training and accelerated convergence in cross-silo settings. ACM Transactions on Intelligent Systems and Technology (TIST)  \textbf{13}(5),  1--29 (2022)
	
	\bibitem{sun2019reallybackdoorfederatedlearning}
	Sun, Z., Kairouz, P., Suresh, A.T., McMahan, H.B.: Can you really backdoor federated learning? (2019), \url{https://arxiv.org/abs/1911.07963}
	
	\bibitem{walter2024mitigating}
	Walter, K., Mohammady, M., Nepal, S., Kanhere, S.S.: Mitigating distributed backdoor attack in federated learning through mode connectivity. In: Proceedings of the 19th ACM Asia Conference on Computer and Communications Security. pp. 1287--1298 (2024)
	
	\bibitem{wang2020attack}
	Wang, H., Sreenivasan, K., Rajput, S., Vishwakarma, H., Agarwal, S., Sohn, J.y., Lee, K., Papailiopoulos, D.: Attack of the tails: Yes, you really can backdoor federated learning. Advances in Neural Information Processing Systems  \textbf{33},  16070--16084 (2020)
	
	\bibitem{FedNova}
	Wang, J., Liu, Q., Liang, H., Joshi, G., Poor, H.V.: Tackling the objective inconsistency problem in heterogeneous federated optimization. arXiv preprint arXiv:2007.07481  (2020), \url{https://arxiv.org/abs/2007.07481}
	
	\bibitem{xiao2017fashionmnistnovelimagedataset}
	Xiao, H., Rasul, K., Vollgraf, R.: Fashion-mnist: a novel image dataset for benchmarking machine learning algorithms (2017), \url{https://arxiv.org/abs/1708.07747}
	
	\bibitem{xie2019dba}
	Xie, C., Huang, K., Chen, P.Y., Li, B.: Dba: Distributed backdoor attacks against federated learning. In: International conference on learning representations (2019)
	
	\bibitem{xu2022more}
	Xu, J., Wang, R., Koffas, S., Liang, K., Picek, S.: More is better (mostly): On the backdoor attacks in federated graph neural networks. In: Proceedings of the 38th Annual Computer Security Applications Conference. pp. 684--698 (2022)
	
	\bibitem{10555053}
	Zhan, D., Hai, R.: Will sharing metadata leak privacy? In: 2024 IEEE 40th International Conference on Data Engineering Workshops (ICDEW). pp. 317--323 (2024). \doi{10.1109/ICDEW61823.2024.00047}
	
	\bibitem{zhang2020batchcrypt}
	Zhang, C., Li, S., Xia, J., Wang, W., Yan, F., Liu, Y.: $\{$BatchCrypt$\}$: Efficient homomorphic encryption for $\{$Cross-Silo$\}$ federated learning. In: 2020 USENIX annual technical conference (USENIX ATC 20). pp. 493--506 (2020)
	
	\bibitem{zhao2018federated}
	Zhao, Y., Li, M., Lai, L., Suda, N., Civin, D., Chandra, V.: Federated learning with non-iid data  (2018). \doi{10.48550/ARXIV.1806.00582}, \url{https://arxiv.org/abs/1806.00582}
	
\end{thebibliography}

\end{document}